%% file: 00-main.tex
\documentclass[twoside,leqno,twocolumn]{article}

\usepackage[letterpaper]{geometry}

\usepackage{balance}
\usepackage{ltexpprt}
\usepackage{hyperref}
\usepackage{cite}
\usepackage{amsmath,amssymb,amsfonts}
\usepackage{mathtools}
\usepackage{graphicx}
\usepackage{textcomp}
\usepackage{xcolor}
\def\BibTeX{{\rm B\kern-.05em{\sc i\kern-.025em b}\kern-.08em
    T\kern-.1667em\lower.7ex\hbox{E}\kern-.125emX}}
\input{command.tex}
\usepackage{tabularx,booktabs}

\begin{document}
\title{Deep Efficient Private Neighbor Generation \\ for Subgraph Federated Learning}
\author{Ke Zhang\thanks{cszhangk@connect.hku.hk, ClusterTech Limited.}
\and Lichao Sun\thanks{lis221@lehigh.edu, Lehigh University.} \and Bolin Ding\thanks{bolin.ding@alibaba-inc.com, Alibaba Group.} \and Siu Ming Yiu\thanks{smyiu@cs.hku.hk, The University of Hong Kong.} \and Carl Yang\thanks{j.carlyang@emory.edu, Emory University.}}

\date{}

\maketitle

% Copyright Statement
% When submitting your final paper to a SIAM proceedings, it is requested that you include
% the appropriate copyright in the footer of the paper.  The copyright added should be
% consistent with the copyright selected on the copyright form submitted with the paper.
% Please note that "20XX" should be changed to the year of the meeting.

% Default Copyright Statement
\fancyfoot[R]{\scriptsize{Copyright \textcopyright\ 2024 by SIAM\\
Unauthorized reproduction of this article is prohibited}}

% Depending on which copyright you agree to when you sign the copyright form, the copyright
% can be changed to one of the following after commenting out the default copyright statement
% above.

%\fancyfoot[R]{\scriptsize{Copyright \textcopyright\ 20XX\\
%Copyright for this paper is retained by authors}}

%\fancyfoot[R]{\scriptsize{Copyright \textcopyright\ 20XX\\
%Copyright retained by principal author's organization}}

%\pagenumbering{arabic}
%\setcounter{page}{1}%Leave this line commented out.

\begin{abstract} \small\baselineskip=9pt Behemoth graphs are often fragmented and separately stored by multiple data owners as distributed subgraphs in many realistic applications. Without harming data privacy, it is natural to consider the \textit{subgraph federated learning} (subgraph FL) scenario, where each local client holds a subgraph of the entire global graph, to obtain globally generalized graph mining models. To overcome the unique challenge of incomplete information propagation on local subgraphs due to missing cross-subgraph neighbors, previous works resort to the augmentation of local neighborhoods through the joint FL of missing neighbor generators and GNNs. Yet their technical designs have profound limitations regarding the utility, efficiency, and privacy goals of FL. In this work, we propose \sage to comprehensively tackle these challenges in subgraph FL. \sage consists of a series of novel technical designs: (1) \underline{D}eep neighbor generation through leveraging the GNN embeddings of potential missing neighbors; (2) \underline{E}fficient pseudo-FL for neighbor generation through embedding prototyping; and (3) \underline{P}rivacy protection through noise-less edge-local-differential-privacy. 
We analyze the correctness and efficiency of \sage, and provide theoretical guarantees on its privacy. 
Empirical results on four real-world datasets justify the clear benefits of proposed techniques.
\end{abstract}


\input{01-intro.tex}
% \input{02-related.tex}
\input{03-problem.tex}

\input{04-method.tex}

\input{05-theory.tex}
\input{06-experiments.tex}

\input{07-conclusion.tex}

\bibliographystyle{plain}
\bibliography{references}

\newpage
\input{08-for_pub_appendix.tex}

\end{document}

%% file: command.tex
\usepackage{graphicx}
\usepackage{algorithm}
\usepackage{subfig}

\usepackage{multirow}
\usepackage{xspace}
\newcommand{\ie}{{\it{i.e.}}}
\newcommand{\eg}{{\it{e.g.}}}

\usepackage{wrapfig}
\newcommand{\F}{F\xspace}

\newcommand{\Ln}{\mathcal{L}^n}

\newcommand{\Lf}{\mathcal{L}^f}

\newcommand{\sage}{FedDEP\xspace}

\newcommand{\gen}{DGen\xspace}
\newcommand{\proto}{Proto\xspace}
\newcommand{\nfdp}{NFDP\xspace}

\newtheorem{mytheorem}{Theorem}[section]
\newtheorem{statement}{Statement}[section]
\usepackage[noend]{algpseudocode}

\usepackage{sidecap}

\usepackage{enumitem}
\newtheorem{innercustomgeneric}{\customgenericname}
\providecommand{\customgenericname}{}
\newcommand{\newcustomtheorem}[2]{%
  \newenvironment{#1}[1]
  {%
   \renewcommand\customgenericname{#2}%
   \renewcommand\theinnercustomgeneric{##1}%
   \innercustomgeneric
  }
  {\endinnercustomgeneric}
}

\newcustomtheorem{customthm}{Theorem}
\newcustomtheorem{customst}{Statement}
\newcustomtheorem{customcor}{Corollary}
\newcustomtheorem{customlemma}{Lemma}
\newcustomtheorem{customdef}{Definition}

%% file: 01-intro.tex
\section{Introduction}
\label{sec:intro}

% \todo{highlight scope: better solve subgraph-level FL}

% \todo{challenges in subgraph-level FL (not just FedSage)}

Graph data mining, one of the most important research domains for knowledge discovery, has been revolutionized by Graph Neural Networks (GNNs), which have established state-of-the-art performance in various domains such as social platforms \cite{kipf2016semi}, e-commerce \cite{he2020lightgcn}, transportation \cite{wang2020traffic}, bioinformatics \cite{xu2019powerful}, and healthcare \cite{choi2017gram}. 
The power of GNNs benefits from training on real-world graphs with millions to billions of nodes and links \cite{ying2018graph}. 
Nowadays, emerging graph data from many realistic applications, such as recommendation, drug discovery, and infectious disease surveillance, are naturally fragmented, forming distributed graphs of multiple ``data silos''. Moreover, due to the increasing concerns about data privacy and regulatory restrictions, directly transferring and sharing local data to construct the entire global graph for GNN training is unrealistic \cite{voigt2017eu}. Please refer to detailed related work in Appendix C.
%Nowadays, emerging graph data from many realistic applications are naturally fragmented and distributively stored, forming  due \cite{li2020federated}. Further due to privacy concerns and regulatory restrictions, the entire graph becomes distributed subgraphs of multiple ``data silos''. Among subgraphs, direct data transfer is forbidden \cite{zheng2020learning, voigt2017eu}, which is commonly seen in medical and healthcare domains, such as AI-based drug discovery, and tracking of infectious diseases.

\begin{figure*}[h]
\centering
\footnotesize
\includegraphics[width=6.3in]{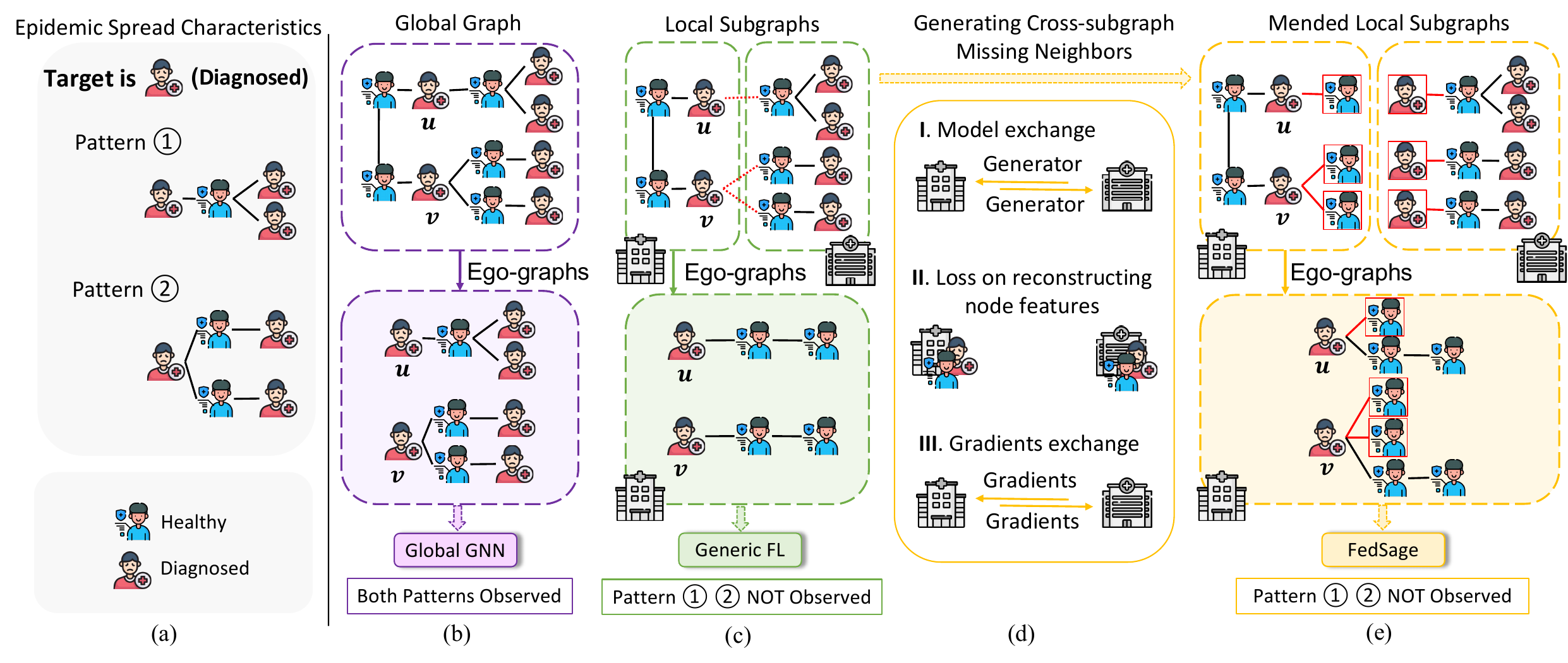}
\caption{A toy example of modeling the spread of infectious disease in a distributed subgraph FL system. The black lines are the close contact relations between people, and the dashed red lines are the cross-subgraph missing links. Red solid lines are the generated links, and the people figures with red solid rectangles are the generated neighbors. (a) The reason for a target to be diagnosed when his/her direct contacts are all healthy can be attributed to Pattern \textcircled{1}: some healthy neighbors directly contact with many diagnosed ones, or Pattern \textcircled{2}: many healthy neighbors directly contacts with diagnosed ones. 
(b) If the global graph is available, both patterns are observable and centralized GNN can correctly identify the reasons for both $u$ and $v$ to be infected. 
(c) In the more realistic setting of local subgraphs, neither of the patterns is observable and GNN obtained from generic FL (such as FedAvg) will fail to learn why $u$ and $v$ are infected. 
(d) FedSage tries to recover 1-hop missing neighbors across local subgraphs through three steps, which require significant extra communication and computation.
(e) Unfortunately, even if all 1-hop missing neighbors can be generated accurately, GNN obtained through FedSage still fail because the correct patterns require access to deeper missing neighbors. %for the cross-subgraph missing neighbor generation in FedSage. Step I and III require a significant amount of data sharing. Step II and III can potentially leak sensitive node features. (e) Mended subgraphs and FedSage of the final classifier. With only $u$ and $v$'s direct contacts with health people reconstructed, the 2-hop ego-graphs of $u$ and $v$ still fail to capture Pattern \textcircled{1} or \textcircled{2}. 
% Such observation is caused by the incomplete context of generated nodes.
}\vspace{-4mm}
\label{fig:system}
\end{figure*}

Federated learning (FL) is a promising paradigm for distributed machine learning that addresses the data isolation problem, which has recently received increasing attention in various applications \cite{yang2019federated}. Despite its successful applications in domains like computer vision \cite{liu2020fedvision} and natural language processing \cite{lin2021fednlp} where data samples (i.e., images or documents) hardly interact with each other, FL over graph data manifests unique challenges due to the complex node dependencies, structural patterns, and feature-link correlations~\cite{yao2022fedgcn, baek2023personalized}.
% In graph learning, data samples (nodes) are correlated by links, which are essential for downstream tasks~\cite{}. However, in FL systems, such correlations can be incomplete or in skew distribution due to the fragmented storage and raw data sharing restrictions~\cite{}. 
% There are two major directions for FL over graph data, federated learning over distributed graphs~\cite{} and federated learning over distributed subgraphs~\cite{}. 
In this work, we focus on one of the most common and challenging scenarios of \textit{federated learning over distributed subgraphs} (subgraph FL), where clients hold subgraphs of largely disjoint sets of nodes and their respective links, as shown in Fig.~\ref{fig:system} (b) and (c). One unique challenge in subgraph FL is the incomplete neighborhood of nodes in the local subgraphs caused by cross-subgraph missing neighbors, that is, potential neighboring nodes captured by other local subgraphs. This phenomenon cannot be properly handled by generic FL mechanisms such as FedAvg for GNN training.
Targeting this limitation, Zhang et al.~propose FedSage~\cite{zhang2021subgraph}, where a \textit{missing neighbor generator} is collaboratively learned across clients to retrieve cross-subgraph missing neighbors and better approximate GNN training on the entire global graph (Fig.~\ref{fig:system} (d))\footnote{For simplicity, we refer to FedSage+ in~\cite{zhang2021subgraph} as FedSage.}. %a novel FL scheme that enables a distributed subgraph system to collaboratively train a generalized node classifier along with a missing neighbor generator to resolve the problem caused by local subgraph disconnection. 
FedSage's success justifies the necessity of completing information in local neighborhoods.
Yet it has several deficiencies in complex and sensitive real-world scenarios, regarding \textit{utility}, \textit{efficiency}, and \textit{privacy}. %as articulated in Fig.~\ref{fig:system} (e). When a city's healthcare system is studying newly emerged infectious epidemics across its hospitals, FedSage can fail to correctly capture all the disease spreading patterns while under significant overheads and privacy concerns. FedSage's limitations lie in \textit{utility}, \textit{efficiency}, and \textit{privacy}, which are the key challenges in the subgraph FL as well.

\noindent\textbf{Limited Utility.} The missing neighbor generator in FedSage can only recover 1-hop missing neighbors and does not further propagate their features to other neighboring nodes. As illustrated in Fig. \ref{fig:system} (e), for predictions where neighbors further than 1 hop away are important, the model will fail. Such limitation is verified by the limited performance gain of FedSage+ over vanilla FedSage without neighbor generator in \cite{zhang2021subgraph} as well as our empirical results in Table \ref{table:main-table}. %regarding the actual information exchanged across the system. The inadequacy of shared information can mislead the final model as it fails to capture complete neighborhoods for local nodes, as described in Fig.~\ref{fig:system} (e). 
% It is also empirically reflected in both~\cite{zhang2021subgraph} and this paper.
        
\noindent\textbf{Significant Overhead.}
The FL training of missing neighbor generators in FedSage incurs substantial inter-client communication costs (Step I in Fig.~\ref{fig:system} (d)), in addition to the standard client-server communication in FL (Step III), and heavy intra-client computations (Step II). Specifically, for each FL iteration, each client needs to broadcast the generated node embeddings to all other clients and receive training gradients for the neighbor generator in Step I, and %model and receive gradients from all other clients in the system. As for local computations in Step II of Fig.~\ref{fig:system} (d), 
each client needs to repeatedly search across its entire node set to find the most likely cross-subgraph missing neighbors in Step III. Other existing studies, such as FedGraph~\cite{chen2021fedgraph} and Lumos\cite{pan2023lumos}, request the central server to manage the FL process with auxiliary data, which induce unneglectable communication and computation overhead.
      
\noindent\textbf{Privacy Concerns.} For subgraph FL, FedSage shares GNN gradients and node embeddings instead of raw data, but without specific privacy protection, the gradients and embeddings directly computed from raw data are prone to privacy leakage, such as %, clients share gradients computed directly on individual node features.
%Plainly sharing such gradients can lead to privacy leakage for local nodes 
via inference attacks \cite{nasr2019comprehensive, luo2021feature} and reconstruction attacks~\cite{zhu2019deep, geiping2020inverting}. 
Unlike FedSage, FedGNN~\cite{wu2022federated} and FedGSL~\cite{zhao2022fedgsl} protect privacy for FL over graphs via noise injection~\cite{abadi2016deep, choi2018guaranteeing}. 
FedGSL lacks discussions on the privacy protection of graph properties, while FedGNN requests an additional authority to guarantee privacy, which is not available in general subgraph FL.
% Hence, it  we discuss which consists of only one server and several clients.

Herein, we propose Subgraph \underline{Fed}erated Learning with \underline{D}eep \underline{E}fficient \underline{P}rivate Neighbor Generation (\sage) to address the unique \underline{utility}, \underline{efficiency}, and \underline{privacy} challenges in the subgraph FL setting. % by solving respective limitations of FedSage. 
% Specifically, we develop a series of techniques including deep neighbor generation, embedding-fused graph convolution, Pseudo-FL training on generators, and noiseless edge-local differential privacy. 
% Without complicated cryptology methods, explicitly perturbing the shared data, or introducing additional characters, our proposed techniques tackle the unique utility, efficiency, and privacy challenges in the subgraph FL system.

% Particularly, in \sage, clients first convert their nodes' unstructured multi-hop local contexts into embeddings, namely the deep neighbors. Then, local clients federally train deep neighbor generators to generate cross-subgraph missing deep neighbors. To accomplish the FL training of the generalized global downstream task model, \ie, a node classifier, we propose a novel embedding-fused graph convolution process to incorporate the generated deep neighbors with original local subgraphs for classification. To further reduce the repeated XXXXX. We argue that the novel \sage overcomes the stated three limitations of FedSage.

\noindent\textbf{Utility-wise: Deep Neighbor Generation and Embedding-fused Graph Convolution (\gen).}
To enhance the modeling of cross-subgraph missing neighbors in the system without exponentially increasing computation and communication overheads, we propose a deep neighbor generator, \gen. It generates missing neighbors in depth, by leveraging GNN embeddings of generated neighbors. The generated embeddings contain information from the target node's multiple hops of neighbors~\cite{zhu2021transfer, tang2022graph} and include richer context beyond single node features generated in FedSage. 
To incorporate the generated deep neighbors into the global GNN classifier training, we propose a novel embedding-fused graph convolution process. %We theoretically justify the correctness and effectiveness of the proposed convolution process.
% instead of creating multiple layers of neighbor generators, which can lead to exponential increments for both computation and communication overhead along the layers, 

\noindent\textbf{Efficiency-wise: Deep Embedding Prototyping and Pseudo-FL (\proto).}
To reduce the intra-client computation, we cluster the local GNN embeddings of nodes in each client to construct sets of missing neighbor prototypes. Instead of repeated exhaustive searches for closest neighbors across a client's entire node set as in FedSage, we can find the closest prototypes across the much smaller prototype sets.
% . With prototyping, each client can effectively summarize different local neighborhoods, so the repeated search can be done 
% instead of going through entire local node sets. 
% Technically, we leverage self-trained clustering \cite{yang2020graph, xie2016unsupervised} to learn the cluster assignments and calculate the prototype embeddings. 
To further reduce the inter-client communication, we propose pseudo-FL by sharing the prototype embeddings across the system before the training of \gen. Thus, clients can conduct closest neighbor searches locally without communications while still achieving FL for \gen. Similarly to \cite{tan2022fedproto}, sharing the prototype embeddings instead of node embeddings can also lead to empirical privacy benefits due to the difficulty in inference attacks from aggregated models.
%We further propose novel Pseudo-FL for clients to obtain FL-trained \gen models on prototype sets without actually sharing data along the FL process. Technically, prototype sets are shared across the system before the FL process. Then, clients can locally search cross-subgraph prototypes to train its generator. In this way, Pseudo-FL sharply reduces the communication to zero with achieving the same effect for federally training \gen.

\noindent\textbf{Privacy-wise: Noise-free differential privacy through random sampling (\nfdp).} We aim to theoretically guarantee rigorous edge-local-differential-privacy (edge-LDP), which protects edges' existence within local node's neighborhoods in distributed subgraphs~\cite{qin2017generating}. Particularly, we achieve noise-free edge-LDP by transferring noise-free differential privacy from general domains \cite{ijcai2021p216} to edge-LDP, without embracing complicated cryptology techniques, explicitly perturbing shared data, or introducing additional roles into the system as previous work~\cite{wu2022federated}. Technically, we incorporate two stages of random sampling into \sage, \ie, (1) mini-batching: random neighborhood sampling in each graph convolution layer \cite{hamilton2017inductive}; and (2) Bernoulli-based generation selection: randomly sampling generated deep neighbors by a Bernoulli sampler in \gen.
% as they bring randomness into their respective model training process 

%We formally analyze the efficiency and provide the theoretical privacy guarantees of FedSage2. 
Extensive experiments on four real-world graph datasets justify the utility, efficiency, and privacy benefits of \sage.
% that compare our proposed techniques with FedSage, FedGNN, several ablations of \sage, and other baselines 

%% file: 03-problem.tex
\section{Problem Formulation}
\label{subsec:problem}

\noindent\textbf{Federated Learning with Distributed Subgraphs.} We denote a global graph as $G=\{V,E,X\}$, where $V$ is the node set, $X$ is the respective node feature set, and $E$ is the edge set. In the subgraph FL system, we have one central server $S$ and $M$ clients with distributed subgraphs. $G_i=\{V_i,E_i,X_i\}$ is the subgraph owned by $D_i$, for $i \in [M]$, where $V = \bigcup_{i=1}^{M} V_i$.
%\paragraph{$\bold{Problem~setup}$} 
%In this system, we assume that the dataless central server $S$ only maintains a graph model, and no direct sharing of nodes nor edges is allowed.
%To simulate the scenario with most information loss, 
For simplicity, we assume no overlapping nodes shared across data owners, \ie, $V_i \cap V_j = \emptyset$ for any $i\neq j \in [M]$. For an edge $e_{v,u}\in E$, where $v\in V_i$ and $u\in V_j$, we have $e_{v,u}\notin E_i\cup E_j$. That is, $e_{v,u}$ might exist in reality but is missing from the whole system.
The system exploits an FL framework to collaboratively learn a global node classifier $\F$ on isolated subgraphs in clients, without raw graph data sharing. 

% The learnable weights $W$ in $\F$ are optimized %for queried ego-graphs 
% following the distribution of %ones drawn from 
% the global graph $G$. We formalize the problem as finding $W^*$ that minimizes the risk 
% %{\small 
% \begin{equation*}
% \label{eq:goal}
%     W^{*} = \argmin_{W}\R(\F(W)) =\argmin_{W}\frac{1}{M}\sum_i^M \R_i(\F_i(W))),
% \end{equation*}
% %}

% where $\R_i(\F_i(W)) \coloneqq \mathbb{E}[\ell(\F_i(W;G_i),Y_i))]$ is the local empirical risk, 
% % and $\ell \coloneqq \frac{1}{|V_i|}\sum_{v\in V_i}l(W;G_i(v),y_v)$ is a task-specific loss function.
% and $\ell(\cdot)$ is a task-specific loss function.}

%% file: 04-method.tex
\section{\sage}
\label{sec:method}

\begin{figure}[h]
\centering
\includegraphics[width=3.2in]{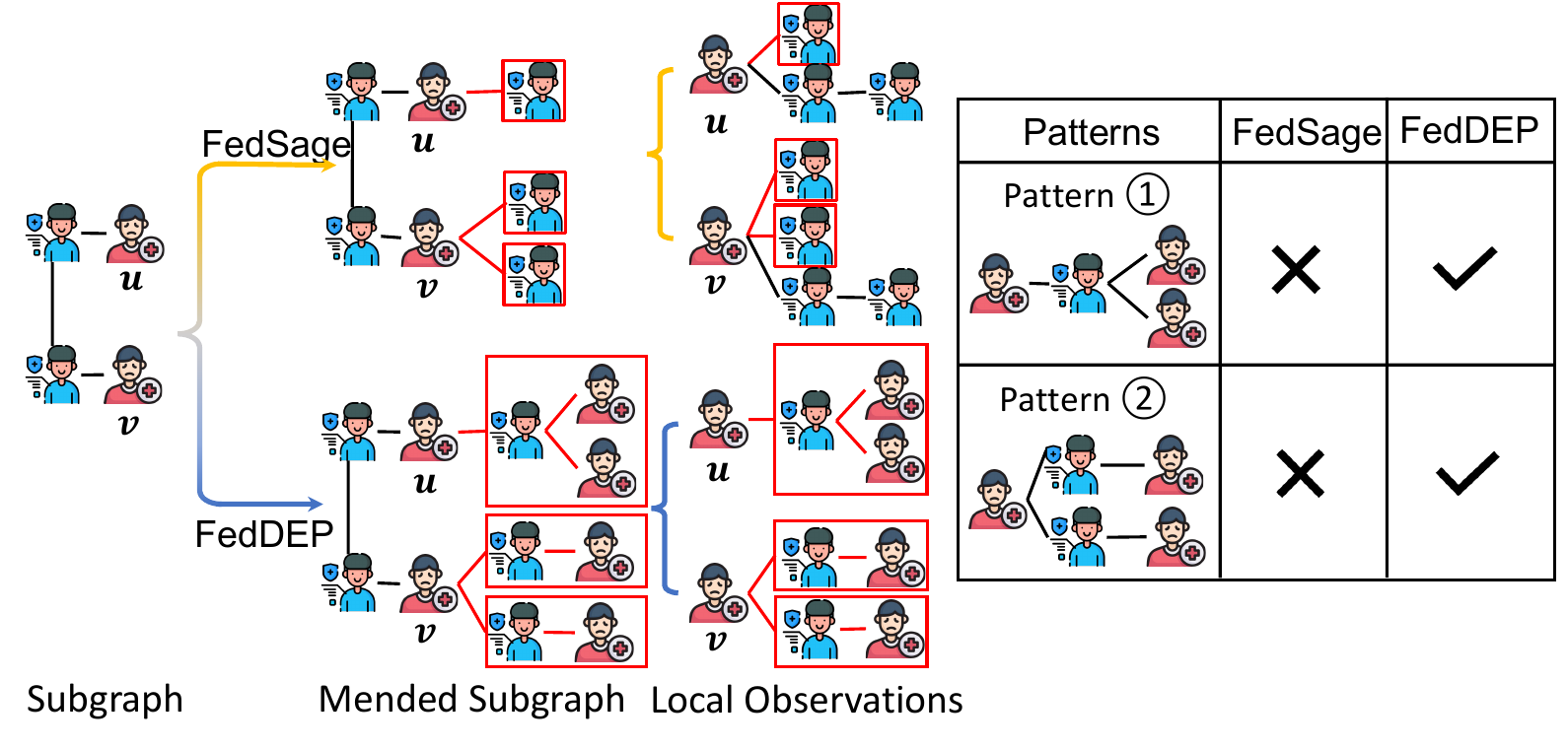}
\caption{Technical motivation of \sage against FedSage. {\normalfont \sage generates information of multiple hops of neighbors to provide the subgraph with richer information for local nodes, compared to the direct missing neighbors generated by FedSage. For more details of FedSage, please refer to the background discussion in Appendix D and the FedSage paper~\cite{zhang2021subgraph}.} \vspace{-15pt}}
\label{fig:sg-utility}
\end{figure}

\begin{figure*}[t]
\centering
\includegraphics[width=0.9\textwidth]{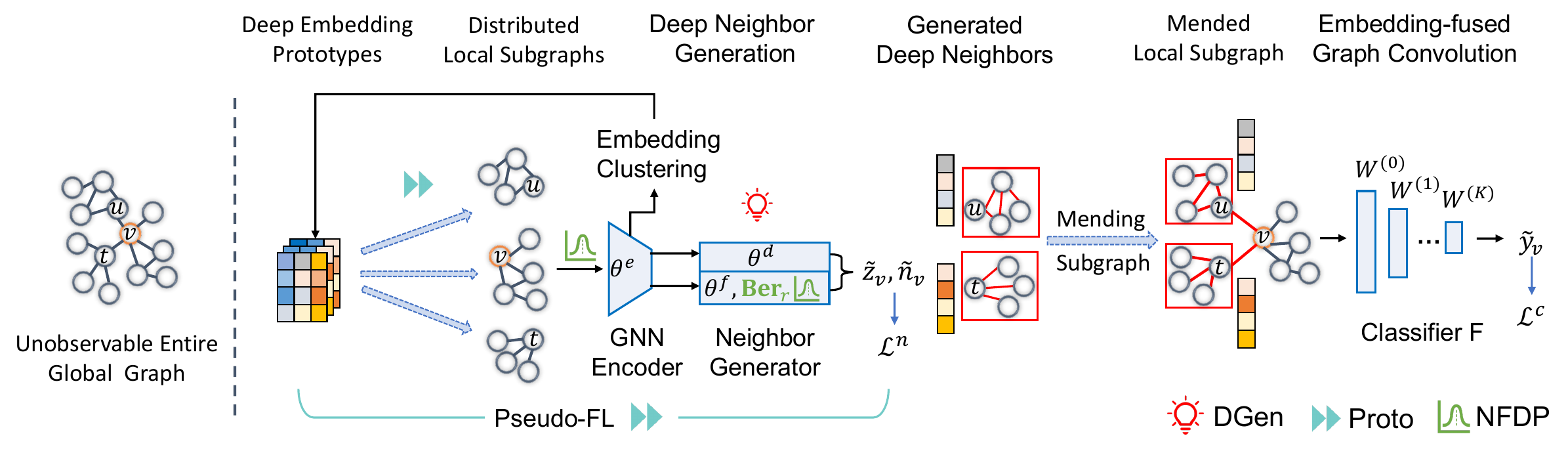}
\caption{Overview of the proposed \sage (with the novel \gen, \proto, and \nfdp components highlighted).} \label{fig:model}
\vspace{-10pt}
\end{figure*}

% \begin{table*}[h]
% \centering
% \small
% \caption{Table of Notations.}\vspace*{-3mm}%The total communication cost is for $E_g$ epochs of training, and $|\theta|$ is $H^g$'s size in FedSage.}
%   \label{table:notation}
% \begin{tabular}{cl}
%    \toprule
%    \centering
%    \small
% Notation & Representation\\
% \midrule
% $H^e,H^d,H^f$ & Different \gen modules.\\
% $\theta^e,\theta^d,\theta^f$ & Parameters of different \gen modules.\\
% $W;W^{(k)}$ & Learnable weights set of proposed embedding-fused GNN; the weights matrix of its $k$-th layer.\\
% $d_x,d_h,d_z,d_y$& Dimensions.\\
% $G_i;G^K_i(v)$ & Local subgraph of $D_i$; its $K$-hop ego-graph of node $v$.\\
% $\tilde{G_i};\tilde{G}^K_i(v)$ & Mended local subgraph of $D_i$; its $K$-hop mended ego-graph of node $v$.\\
% $\tilde{E}_i,\tilde{Z}_i$&
% Mended links and corresponding node embeddings for $D_i$.\\
% $\tilde{n}_v,\tilde{z}_v;$& The predicted number and corresponding embeddings of node $v$'s inter-subgraph missing neighbors;\\
% $\tilde{z}_v^p$& the $p$-th predicted neighbor's embedding for node $v$.\\
% $Z_i;z_v$& Node embedding set of $D_i$; node embedding of node $v$.\\
% $Z'_i;z'_u$& Node embedding prototype set of $D_i$; prototype of node embedding cluster $u$.\\
% $x^k_v$& Node representation of node $v$ at layer $k$.\\
% $\mathcal{L};\Ln_i,\Lc_i$& Loss of proposed \sage; cross-subgraph prototype reconstruction loss and cross-entropy classification loss on $D_i$.\\
% \bottomrule
% \end{tabular}
% \end{table*}

\subsection{Utility Elevation through \gen}
\label{subsec:gen}
The demanding utility challenge left by FedSage is how to further enrich local node contexts regarding deeper missing neighbors. However, directly generating deeper neighbors incurs exponential increments for intra-client computation and inter-client communication. 
To solve this, we propose to leverage the GNN encoder and generate deep embeddings of missing neighbors that captures their multi-hop local contexts in the corresponding subgraphs. Fig.~\ref{fig:sg-utility} illustrates this technical motivation, regarding the toy example in Fig.~\ref{fig:system}.

\noindent\textbf{Deep Neighbor Generation.} Inspired by the design of NeighGen in FedSage, we propose a deep neighbor generator \gen, whose architecture is in the middle of Fig.~\ref{fig:model}. $\theta^e$ and $\{\theta^d,\theta^f\}$ are the learnable parameters of \gen's two components, \ie, the GNN encoder and the embedding generator, respectively.

Unlike NeighGen that generates node features of missing neighbors, \gen generates node embeddings. Particularly, for a node $v$ on graph $G_i$, we have its generated missing deep neighbors as 
\vspace{-2mm}
{\scriptsize
\begin{equation*}
\begin{aligned}
\tilde{n}_v&=\sigma(\theta^{d\top}\cdot Enc(G_i^K(v);\theta^e)),\\
\tilde{z}_v&= Ber_r\left(\sigma\left(\theta^{f\top}\cdot[Enc(G_i^K(v);\theta^e)+\mathbf{N}(0,1)]\right), \tilde{n}_v\right),\vspace{-9pt}
\end{aligned}
\end{equation*}}\vspace{-2mm}

where $Enc$ is the GNN encoder of \gen, $\tilde{n}_v\in\mathbb{N}$, $\tilde{z}_v\in\mathbb{R}^{\tilde{n}_v\times d_z}$, and $d_z$ is the dimension of node embeddings. $Ber_r(a,b)$ is a Bernoulli sampler that independently samples $b$ records from $a$ following $Ber(r)$, with $r$ as a constant. 

In this process, based on the original neighborhood of $v$, \gen first predicts $\tilde{n}_v$, the number of its missing neighbors $n_v$, and then samples $\tilde{n}_v$ embedding vectors for them. Sampling allows the whole process to be generative and trained through variational inference for enhanced robustness \cite{kipf2016variational}. %Ber_r(a,b)$ incorporates the generation with rigorous privacy protection via introducing randomness into it.

\noindent\textbf{Embedding-fused Graph Convolution.} On mending local subgraphs with generated deep neighbors, \ie, attaching nodes with deep embeddings, every client obtains its mended local subgraph $\tilde{G}_i=\{V_i,\tilde{E}_i,X_i,\tilde{Z}_i\}$. 
Yet existing GNNs with vanilla graph convolution process is incapable to incorporate node features with deep embeddings, due to the difference between their feature spaces. To conduct node classification on $\tilde{G}_i$ to fulfill the goal of obtaining a node classifier $\F$, we propose the embedding-fused graph convolution mechanism. 

\noindent\textbf{Downstream~Classifier~\F.}
$\F$ is a model of $K$-layer embedding-fused graph convolution. 

For $v\in V$, $\F$ integrates $v$'s $K$-hop mended ego-graph $\tilde{G}^K_i(v)$ on $\tilde{G}_i$. The convolution is achieved by learnable weights $W$=$\{W^{(k)}|k\in[0,K]\}$, where $W^{(0)}\in\mathbb{R}^{d_h\times (d_x+d_z)}$, for $k\in[K$-$1]$, $W^{(k)}\in\mathbb{R}^{d_h\times (d_h+d_z)}$, and $W^{(K)}\in\mathbb{R}^{d_y\times (d_h+d_z)}$. $d_h$ is the dimension of hidden representation $x^{k}_v$, for node $v$ at layer $k\in[0,K$-$1]$.

For a node $v\in V_i$, its initial representation $x^0_v$ is 
{\scriptsize
\begin{equation*}
\label{eq:emb-sage-init}
    \begin{split}
    x^0_v =\sigma & \left( W^{(0)} \times [x_v|| Agg(\tilde{z}_v)]^\top\right)^\top,
    \end{split}
\end{equation*}
}
where $Agg$ is specified as a mean aggregator, and $||$ is the concatenating function. 

For layer $k\in[K]$, the model computes $v$'s representation $x^k_v$ as
{\scriptsize
\begin{equation*}
\label{eq:emb-sage}
    \begin{split}
    x^k_v =\sigma & \left( W^{(k)} \times [Agg(\{x^{k-1}_u| u\in G^1_i(v)\})^\top|| Agg(\tilde{z}_v)]^\top\right)^\top.
    \end{split}
\end{equation*}
}

After $K$ layers, $\F$ outputs the inference label as $\tilde{y}_v= \text{Softmax}(x^K_v)$.
We formally analyze the correctness of embedding-fused graph convolution and whose proof is in Appendix A.1.
% \footnote{Appendix accessible in https://anonymous.4open.science/r/FedDEP-6F08/.}
\begin{statement}
% [Correctness of embedding-fused graph convolution] 
For a node $v$, at each layer of embedding-fused graph convolution, it aggregates nodes on the impaired ego-graph with the corresponding mended deep neighbor embeddings with separate learnable weights.
\label{thm:correct}
\end{statement}

\noindent\textbf{Inference.} For a node $v$ on the global graph $G$ without mended deep neighbors, $\F$ predicts its label by setting all $\tilde{z}_v$'s to zero vectors.
We justify that $\F$ with $K$ layers of embedding-fused graph convolution in aggregating real neighbors and generated deep neighbors of $L$-hop local contexts, has a similar capacity as the original GCN in aggregating an ($K$+$L$)-hop ego-graph as shown below. The proof is provided in Appendix A.2. \vspace{-2mm}

\begin{statement}
%[Comparison between embedding-fused graph convolution and original graph convolution] 
For a node $v$, we denote the prediction, computed by $K$ layers of embedding-fused graph convolution on its $K$-hop impaired ego-graph mended with deep neighbors of $L$-hop, as $\tilde{y}'_v$, and the prediction, computed by ($K$+$L$) layers of graph convolution on its ($K$+$L$)-hop ego-graph, as $\tilde{y}_v$, where $K,L\in \mathbb{N}^*$. $\tilde{y}'_v$ and $\tilde{y}_v$ encode the same local context of $v$.
\label{thm:convolve}
\end{statement}

% To better serve the downstream classification task, \gen is jointly trained with $\F$ by minimizing $\Ln_i+\Lc_i$, where $\Lc_i$ is the cross entropy function as
% \begin{equation}\label{eq:loss-graphsage}
%    \Lc_i=l(W|\tilde{G}^K_i(v),y_v) = -\left[y_{v} \log \tilde{y}_{v}+\left(1-y_{v}\right) \log \left(1-\tilde{y}_{v}\right)\right].
% \end{equation}

% To this end, the generated deep neighbors from \gen provide assistance for $\F$ in understanding the original node contexts, while $\F$ supplies \gen with additional supervised information. Thus, the joint model can better serve the downstream task. 

\noindent\textbf{FL for the Joint Model of \gen and $\F$.}
\label{subsubsec:FL-joint}
We follow FedSage to jointly train \gen and $\F$ via FL with
    {\scriptsize
\begin{equation*}
\begin{aligned}
  \Ln_{i,i}= \frac{1}{|\bar{V}_i|} \sum_{v\in \bar{V}_i}& [\lambda^d L_1^S(\tilde{n}_v-n_v)
 % \right.\\
    % &\left.
    +\lambda^f \sum_{p\in[\tilde{n}_v]}
	\min_{u\in \bar{\mathcal{N}}_{i}(v)}(||\tilde{z}_{v}^{p}-z_{u}||^2_2)],
\end{aligned}
\end{equation*}}
where $\tilde{z}^p_v$ is the $p$-th generated embeddings in $\tilde{z}_v$, $n_v$ and $z_u$ are the local ground truths retrieved from the hidden information.
Within the joint model, \gen can enrich local nodes' neighborhoods to approximate the complete ones on the unobservable global graph, for $\F$ to conduct accurate and generalized node classification, while $\F$ can supply \gen with task-oriented supervision.

\subsection{Efficiency Elevation through \proto}
When the system conducts FL over the joint model of \gen and $\F$, it encounters significant overhead regarding %although only exchanging the generated deep neighbors avoids the transmissions over generators in FedSage,  exhaustive search space for 
intra-client computations for closest potential neighbor search and inter-client communications for frequent gradient/embedding exchange.
To fundamentally reduce both costs, we propose pseudo-FL with deep neighbor prototype generation. We term it as \proto.

\noindent\textbf{Deep Neighbor Prototype Generation.} 
% In the deep neighbor prototype generation, \gen generates prototypes of node embedding clutters, instead of embeddings or features of individual nodes. 
Technically, every client $D_i$ first locally trains a GNN of the same construction as the final \F. Next, $D_i$ retrieves pre-computed local node embeddings $Z_i$ from the local GNN, and groups them into $C$ clusters by a clustering function. Then $D_i$ obtains its prototype set as $Z'_i=\{mean(z_v|v\in V_i, z_v\text{ in cluster }c)|c\in[C]\}$. 

Subsequently, the cross-subgraph prototype reconstruction loss $\Ln_{i}$ is computed on the prototype sets as

{\scriptsize
\begin{equation*}
\begin{aligned}
\mathcal{L}^{n}_{i} = \frac{1}{|\bar{V}_i|} \sum_{v\in\bar{V}_i} [&\beta^d L_1^S(\tilde{n}_v-n_v) +\beta^f\sum_{p\in[\tilde{n}_v]}
	\min_{u\in \bar{\mathcal{N}}_{i},z'_{u}\in Z'_i}(||\tilde{z}_{v}^{p}-z'_{u}||^2_2)\\
&+\beta^n\sum_{j\in[M]\setminus\{i\}}\sum_{p\in[\tilde{n}_{v}]}
	\min_{z'_{u}\in Z'_j} (||\tilde{z}_{v}^{p}-z'_{u}||^2_2)],
\end{aligned}
\label{eq:sage_gen}
\end{equation*}
}
where $\beta$'s are constants.

In this way, the intra-client search space in computing the reconstruction loss reduces from $|V|$ to $M*C$ prototypes.

The FL training of the joint model with prototyping is to minimize the following objective function
{\scriptsize
\begin{equation}
\mathcal{L}=\frac{1}{M}\sum_{i\in[M]}\mathcal{L}_{i}=\frac{1}{M}\sum_{i\in[M]}(\mathcal{L}^{n}_{i} + \mathcal{L}_i^{c}),
\label{eq:sage}
\end{equation}
}
where $\mathcal{L}_i^{c}$ is the cross-entropy loss computed on deep neighbor prototype mended subgraph.

\noindent\textbf{Pseudo-FL with Cross-subgraph Prototype Generation.} To further reduce communication costs without forbidding clients from learning across the system, we propose pseudo-FL based on \proto.
% tIt enables \gen to learn cross-subgraph prototypes without actually sharing training-related data during the FL process, especially in computing $\mathcal{L}^{n}_{i}$ with Eq.~\eqref{eq:sage_gen}.
In pseudo-FL, each $D_i$ sends $Z'_i$ across the system before the FL process. For every $D_i$, after obtaining $Z'=\{Z'_j|j\in[M]\}$, it can conduct the FL process for \gen by \textit{locally} computing the %cross-subgraph deep neighbor prototype generation loss $\mathcal{L}^{n}_{i}$  % required, $D_i$ enables its \gen to generate cross-subgraph deep neighbor prototypes by \textit{locally} computing the 
cross-subgraph deep neighbor reconstruction loss $\mathcal{L}_i$ in Eq.~\eqref{eq:sage} with zero inter-client communication.
Then, among deep neighbor prototype mended subgraphs, the system conducts FL (\eg, FedAvg) to attain the final classifier by minimizing $\mathcal{L}$ in Eq.~\eqref{eq:sage}.

%We term the entire process as \sage, subgraph FL with deep efficient private neighbor generation. The entire training process is shown in Fig.~\ref{fig:model}.

{\setlength{\parindent}{0cm}\textbf{Efficiency Analysis.}} The main difference between FedSage, \sage and generic FL frameworks (\eg, FedAvg) is the additional learning of neighbor generators. We analyze the additional overhead caused by the FL training of the neighbor generators for three different frameworks in Table~\ref{table:efficiency}.

The computation complexity for \sage is decreased from FedSage by the reduction in the generated dimension (for real-world datasets, $d_x$ can be a few thousand~\cite{shchur2018pitfalls}, while $d_z$ in \sage is often less than 300). By prototyping deep neighbors into $C$ clusters, where $C$ can be rather small such as 10, the computation complexity of \sage is further significantly decreased.
Communication-wise, the cost of FedSage is dominated by generator's size $|\theta|$, which can be as large as 3MB even for a simple two-layer GCN model. \sage without \proto (\sage$_\text{/\proto}$) reduces the cost by sharing deep neighbors. With pseudo-FL, \sage cuts the communication to zero by sharing $O(MCd_z)$ data ahead of training \gen. \vspace{-2mm}

\begin{table}[h]
\centering
\small
\scriptsize
\caption{The additional overhead caused by the FL training of neighbor generator (one round of updating a generator for one node). }%The total communication cost is for $E_g$ epochs of training, and $|\theta|$ is $H^g$'s size in FedSage.}
  \label{table:efficiency}
\begin{tabular}{cccc}
   \toprule
   \centering
{FL scheme} & {comp.} & {comm./epoch} & {total comm.}\\
\midrule
{FedSage} &  {$O(|V|\tilde{n}_v d_x)$} & {$O(M|\theta||h_v^K|)$} & {$O(E_gM|\theta||h_v^K|)$}\\
{\sage$_\text{/\proto}$} & {$O(|V|\tilde{n}_v d_z)$} & {$O(M\tilde{n}_vd_z)$} & {$O(E_gM\tilde{n}_vd_z)$}\\
{\sage} & {$O(MC^2 d_z)$} & 0 & {$O(MCd_z)$} \\
\bottomrule
\end{tabular}
\end{table}

%% file: 05-theory.tex
\subsection{Privacy Guarantees through \nfdp}
\label{subsec:privacy}
We theoretically analyze the edge-LDP property of \sage achieved by our novel noise-free DP mechanism (\nfdp). Even without explicitly injecting random noises into the original local neighborhoods, our proposed framework sustains strong privacy protections by rigorous edge-LDP.

\begin{mytheorem}[Noise-free edge-LDP of \sage] 
For a distributed subgraph system, on each subgraph, given every node's $L$-hop ego-graph with its every $L$-1 hop neighbors of degrees by at least $D$, \sage unifies all subgraphs in the system to federally train a joint model of a classifier and a cross-subgraph deep neighbor generator. By learning from deep neighbor embeddings that are obtained from locally trained GNNs in $N$ epochs of mini-batch training with a sampling size for each hop as $d$, \sage achieves $(\log(1+r(e^{\tilde{\varepsilon}}\text{-}1),r\tilde{\delta})$-edge-LDP, where
{\scriptsize
\begin{equation*}
\begin{aligned}
\tilde{\varepsilon}&=\min \{LN \varepsilon, LN \varepsilon\frac{(e^{\varepsilon}-1)}{e^{\varepsilon}+1}+\varepsilon U\sqrt{2 LN} \},\\
\tilde{\delta}&=(1-\delta)^{LN}(1-\delta'), \quad \delta'\in[0,1],
\end{aligned}
\end{equation*}
}
and $U$ = {\scriptsize$\min \{\sqrt{\ln (e+\frac{\varepsilon\sqrt{LN }}{\delta'})}, \sqrt{\ln (\frac{1}{\delta'})}\}$}. $r$ is the expected value of the Bernoulli sampler in \gen. When $d$<$D$, $(\varepsilon,\delta)$ are tighter than {\scriptsize$(\ln\frac{D+1}{D+1-d}, \frac{d}{D})$}; when $d\geq D$, $(\varepsilon,\delta)$ are tighter than {\scriptsize$(d\ln\frac{D+1}{D}, 1-(\frac{D-1}{D})^d)$}. Both pairs of $(\varepsilon,\delta)$ serve as the lower bounds of the edge-LDP protection under the corresponding cases.
\label{thm:sage}
\end{mytheorem}

Since both $\epsilon$ and $\delta$ are simultaneously affected by the sampling size of local model training, for simplicity, we choose $\epsilon$ to evaluate privacy costs in our experiments.
The proof of the Theorem follows general noise-free DP \cite{ijcai2021p216}, the rule of the composition of DP mechanisms \cite{kairouz2015composition}, and privacy amplification by subsampling \cite{balle2020privacy}.
Due to the space limit, the detailed proof is in Appendix B.

% \begin{remark}[Further tighten $\tilde{\varepsilon}$ and $\tilde{\delta}$ via pre-sampling]

% \end{remark}

{\setlength{\parindent}{0cm}\textbf{Discussions.}} \proto does not theoretically tighten the privacy bound of edge-LDP. %Hence, under the same conditions in Theorem~\ref{thm:sage}, \sage without \proto is of the same edge-LDP as \sage.
% As prototyping does not include additional random process, 
However, unlike individual node features or node embeddings, prototypes in \proto are aggregated data and do not have a one-to-one correspondence with individual nodes. Thus, \proto not only benefits FL efficiency, but also enhances the empirical privacy protection of \sage. % the original embeddings and corresponding neighborhoods from being recovered, compared to \sage without \proto.

%% file: 06-experiments.tex
\section{Experiments}
\label{sec:exp}

\begin{table*}[t]
  \caption{Node classification results. The top two models are highlighted (except for Global).}\vspace*{-2mm}
  \label{table:main-table}
  \centering
  \footnotesize
  \begin{tabular}{ccccccc}
   \toprule
   
   Training&M=3&M=5&M=10&M=3&M=5&M=10\\
   \cmidrule(r){2-4}\cmidrule(r){5-7}
   Frameworks&\multicolumn{3}{c}{\textbf{Cora}$\quad Global$: 0.8955$\pm$.004 }&\multicolumn{3}{c}{\textbf{CiteSeer}$\quad Global$: 0.7741$\pm$.005}\\
   \cmidrule(r){1-1}\cmidrule(r){2-4}\cmidrule(r){5-7}
    Local&0.5776$\pm$.025&0.4486$\pm$.108&0.4334$\pm$.083&{0.6541$\pm$.028}&{0.5802$\pm$.056}&{0.4200$\pm$.110}\\
FedAvg &0.8571$\pm$.015&0.8555$\pm$.014&0.8528$\pm$.020&{0.7646$\pm$.011}&{0.7496$\pm$.010}&{0.7350$\pm$.008}\\
 FedGNN &0.8823$\pm$.017&0.8670$\pm$.010&0.8675$\pm$.011	&{0.7850$\pm$.013}&{0.7927$\pm$.014}&0.7823$\pm$.011\\ 
FedGraph & 0.8693$\pm$.002 & 0.8602$\pm$.004 & 0.8507$\pm$.010 &  0.7720$\pm$.033 & 0.7834$\pm$.021 & 0.7633$\pm$.012\\ 
FedGSL & 0.8633$\pm$.001 & 0.8620$\pm$.006 & 0.8613 $\pm$.040 &  0.7810$\pm$.043 & 0.7900$\pm$.014 & 0.7567$\pm$.011\\ 
FedSage&0.8708$\pm$.009&0.8790$\pm$.009&0.8588$\pm$.010&{0.7818$\pm$.002}&{0.7805$\pm$.017}&{0.7656$\pm$.010}\\
\cmidrule(r){1-1}\cmidrule(r){2-4}\cmidrule(r){5-7}							
\sage$_{\text{w/o \gen}}$& 0.8718$\pm$.007& 0.8717$\pm$.004& 0.8583$\pm$.007	&{0.7616$\pm$.006}&{0.7806$\pm$.005 }&0.7413$\pm$.005\\
\sage$_{\text{w/o \proto}}$ &{\bf0.8911$\pm$.003}&{\bf0.8900$\pm$.003}&{\bf0.8916$\pm$.017}&{\bf0.8107$\pm$.018}&{\bf0.7995$\pm$.009}&{\bf0.8080$\pm$.010}\\  
   \sage$_{\text{w/o \nfdp}}$&0.8883$\pm$.018&0.8703$\pm$.015&0.8747$\pm$.009	&{0.7846$\pm$.018}&{0.7913$\pm$.011}&0.7882$\pm$.016	\\ 
   
 \sage&{\bf0.8894$\pm$.016}&{\bf0.8883$\pm$.011}&{\bf0.8801$\pm$.009}& {\bf0.7927$\pm$.014} &{\bf0.7940$\pm$.016}&{\bf0.8040$\pm$.022}\\
     
   %   \cmidrule(r){1-1}\cmidrule(r){2-4}\cmidrule(r){5-7}

   % Global&\multicolumn{3}{c}{0.8955$\pm$.004}&\multicolumn{3}{c}{0.7741$\pm$.005}\\
    % \bottomrule
    \toprule
   &\multicolumn{3}{c}{\textbf{PubMed}$\quad Global$: 0.8996$\pm$.001}&\multicolumn{3}{c}{\textbf{MSAcademic}$\quad Global$: 0.9597$\pm$.001}\\
   % \cmidrule(r){2-4}\cmidrule(r){5-7}
   % Model&M=3&M=5&M=10&M=3&M=5&M=10\\
    % \midrule
    \cmidrule(r){1-1}\cmidrule(r){2-4}\cmidrule(r){5-7}
    Local&0.8287$\pm$.008&0.7879$\pm$.032&0.4364$\pm$.112	&{0.7906$\pm$.011}&0.7713$\pm$.099&0.5445$\pm$.072	\\		
  FedAvg&0.7149$\pm$.012&0.7260$\pm$.002&0.6954$\pm$.026&0.6986$
\pm$0.002&{0.6908$\pm$.020}&{0.6705$\pm$.012}\\
  FedGNN & 0.9009$\pm$.006&0.8854$\pm$.005	&0.8576$\pm$.006&{0.9403$\pm$.002}&0.9396$\pm$.001&{\bf0.9362$\pm$.001}\\ 
   FedGraph & 0.8921$\pm$.002& 0.8774$\pm$.004& 0.8581$\pm$.007&  0.9311$\pm$.002& 0.9225$\pm$.007    & 0.9244$\pm$.005 \\
   FedGSL & 0.8991$\pm$.003 & 0.8815$\pm$.002 & 0.8592 $\pm$.004 &  0.9385$\pm$.003 & 0.93060$\pm$.004 & 0.9268$\pm$.001\\ 
   FedSage&{0.8877$\pm$.008}&0.8794$\pm$.003&{0.8639$\pm$.008}&{0.9359$\pm$.001}&{0.9414$\pm$.001}&{0.9314$\pm$.001}\\

		\cmidrule(r){1-1}\cmidrule(r){2-4}\cmidrule(r){5-7}
\sage$_{\text{w/o \gen}}$& 0.8440$\pm$.001 & 0.8553$\pm$.008 & 0.8273$\pm$.010	&{ 0.9434$\pm$.001 }&{ 0.9416$\pm$.001 }& 0.9331$\pm$.001 \\ 
\sage$_{\text{w/o \proto}}$&{\bf0.9090$\pm$.005}&{\bf0.8885$\pm$.002}&{\bf0.8697$\pm$.004}&{\bf0.9504$\pm$.004}&{\bf0.9455$\pm$.001}&{\bf0.9362$\pm$.001}\\								
  \sage$_{\text{w/o \nfdp}}$&{0.9020$\pm$.007}&0.8819$\pm$.001&0.8605$\pm$.002	&{0.9406$\pm$.001}&0.9387$\pm$.001&0.9339$\pm$.001\\ 
    \sage&{\bf0.9039$\pm$.007}&{\bf0.8872$\pm$.003}&{\bf0.8662$\pm$.003}&{\bf0.9452$\pm$.001}&{\bf0.9422$\pm$.001 }&{0.9351$\pm$.002}\\
   %   \cmidrule(r){1-1}\cmidrule(r){2-4}\cmidrule(r){5-7}
   % Global&\multicolumn{3}{c}{0.8996$\pm$.001}&\multicolumn{3}{c}{0.9597$\pm$.001}\\
    \bottomrule
  \end{tabular}
\end{table*}

We conduct experiments on four real-world graph datasets to verify the benefits of \sage under different scenarios, with in-depth component studies for \gen, \proto, and \nfdp. %We further show case studies in visualizing the training curves of \sage and compared baselines.  

\vspace{-2mm}
\subsection{Experimental Settings}

We synthesize the distributed subgraph system with four widely used graph datasets, \ie, Cora~\cite{sen2008collective}, CiteSeer~\cite{sen2008collective}, PubMed \cite{namata2012query}, and MSAcademic~\cite{shchur2018pitfalls}. 
We follow FedSage \cite{zhang2021subgraph} to synthesize the distributed subgraph system using the Louvain Algorithm~\cite{blondel2008fast}. We split every dataset into 3, 5, and 10 subgraphs of similar sizes, and due to the space limit, whose statistics are presented in Appendix E.
% ,\todo{should we specify this?} note that the number of nodes and edges in this work are different from the ones in FedSage.

We specify the GNN as a two-layer GraphSage with mean aggregator \cite{hamilton2017inductive} and neighbor size 5. Batch size and training epochs are set to 32 and 50. Same parameters are used for \F with embedding-fused graph convolution. The train-val-test ratio is 60\%-20\%-20\% and all loss weights are set to 1. The graph impairing ratio $h$ is set to 0.5. SGD optimization is applied with 0.1 learning rate. $d_z$ is 128 for Cora, 64 for CiteSeer, 256 for both PubMed and MSAcademic, based on the grid search over \{64,128,256\}. We implement \sage on the FederatedScope platform~\cite{wang2022federatedscope} in Python. All experiments are on a server with 8 NVIDIA GeForce GTX 1080 Ti GPUs.\footnote{Code: https://anonymous.4open.science/r/FedDEP-6F08/.}
 
We conduct comprehensive performance evaluations of \sage by comparing following baselines and ablations: 
(1) Global: A GraphSage model trained on the entire global graph without missing links (providing the performance upper bound); 
(2) Local: A set of GraphSage models trained on individual subgraphs; 
(3) FedAvg (\sage without \gen/\proto/\nfdp): A GraphSage model trained across subgraphs by FedAvg;
(4) FedGNN: A GraphSage model trained across subgraphs by FedGNN \cite{wu2022federated};
(5) FedGraph: A GraphSage model trained across subgraphs by FedGraph \cite{chen2021fedgraph};
(6) FedGSL: A GraphSage model trained across subgraphs by FedGSL~\cite{zhao2022fedgsl};
(7) FedSage: A GraphSage model trained across subgraphs by FedSage+~\cite{zhang2021subgraph};
(8) \sage$_{\text{w/o \gen}}$: \sage without deep neighbor generation; %FedSage+~\cite{zhang2021subgraph} trained on missing neighbor feature prototypes
(9) \sage$_{\text{w/o \proto}}$: \sage without pseudo-FL or embedding prototype;
(10) \sage$_{\text{w/o \nfdp}}$: \sage trained with DPSGD instead of noise-free edge LDP;
and (11) \sage: The full \sage model with \gen, \proto and \nfdp.

Cluster numbers in \sage are chosen by grid search over $C\in\{3,5,10,15,20\}$ and will be studied in Section~\ref{subsec:exp_proto}. For FedGNN and \sage$_{\text{w/o \nfdp}}$, we fix their $\sigma$ as 2.0 to achieve the same level of edge-LDP protection as other variations of \sage. We provide results with different privacy budgets in Section \ref{subsec:exp_dp}.% as \sage and \sage w/o proto when $d$=5 and $D$=10.
 
The metric we use is node classification accuracy on the queries sampled from test nodes on the global graph. The reported average accuracy is over three random repetitions. For locally trained models, the scores are further averaged across local models. The corresponding standard deviations are also provided.

\vspace{-2mm}
\subsection{Overall Performances} 

We conduct comprehensive ablation experiments to verify the significant elevations brought by our proposed techniques, as shown in Table~\ref{table:main-table}. 
The most exciting observation is that besides outperforming local models by an average of 27.13\%, \sage distinctly elevates the performance of FedGNN by at most 2.13\%, and FedSage by at most 3.84\%, even by requiring zero communication during the FL of the generators. Notably, similar to FedSage, \sage exhibits its capacities in elevating beyond the global classifier which is supposed to provide the performance upper bound, possibly due to the additional model robustness brought by the missing neighbor generators. As shown in the results of CiteSeer in Table~\ref{table:main-table}, \sage even excels the global model by at most 2.99\%.

Experimental results of comparing \sage with FedSage and \sage$_{\text{w/o \gen}}$ justify the necessity of generating multi-hop cross-subgraph neighbors. Specifically, \sage exceeds FedSage by 1.27\% on average, and \gen improves \sage by 2.49\%. Though \proto can cause slight accuracy loss, \ie, 0.52\% on average between \sage and \sage$_{\text{w/o \proto}}$, it both benefits the efficiency (as shown in Fig. \ref{fig:curves}) and reduces the empirical risks of privacy leakage \cite{tan2022fedproto}.

% Compared to vanilla FedAvg (\sage w/o \gen, w/o prototyping), \sage and \sage w/o prototyping bring respective averaged improvement by 12.79\%, 13.16\%, and averaged 1.41\%, 1.78\% improvements compared to FedSage. 

It is obvious that the more missing links, i.e., more missing information, in the system, the more likely a larger performance elevation from \gen can be brought to vanilla FedAvg and \sage$_{\text{w/o \gen}}$. For the MSAcademic dataset with 10 clients, we infer the reason of FedGNN slightly exceeding \sage to be the significant amount of missing inter-subgraph links (32.94\%). In this circumstance, when \sage further abstracts shared information through \proto, performance degeneration can be the result. Even in this difficult scenario, the generation of prototyped deep neighbors can still help \sage to clearly outperform FedAVG and FedSage. 

Under similar privacy protection, \sage on average exceeds FedGNN and \sage$_{\text{w/o \nfdp}}$ by 0.76\% and 0.61\%, respectively. Without \proto, \sage$_{\text{w/o \proto}}$ on average outperforms FedGNN and \sage$_{\text{w/o \nfdp}}$ by 1.28\% and 1.13\%, respectively. Such gaps justify the advantageous privacy-utility trade-off of our novel \nfdp over DPSGD with noise injection.

\begin{table*}[t!]
  \caption{Component study for \proto with varying cluster numbers $C$ on four datasets with different $M$'s.}
  \label{table:cluster}
  \centering
  \small
  \footnotesize
  \begin{tabular}{ccccccc}
   \toprule
   
   Training&M=3&M=5&M=10&M=3&M=5&M=10\\
   \cmidrule(r){2-4}\cmidrule(r){5-7}
   Frameworks&\multicolumn{3}{c}{\textbf{Cora}$\quad |Y|$=7}&\multicolumn{3}{c}{\textbf{CiteSeer}$\quad |Y|$=6}\\
   \cmidrule(r){1-1}\cmidrule(r){2-4}\cmidrule(r){5-7}
    FedAvg &0.8571$\pm$.015&0.8555$\pm$.014&0.8528$\pm$.020&{0.7646$\pm$.011}&{0.7496$\pm$.010}&{0.7350$\pm$.008}\\		
\sage$_{\text{w/o \proto}}$ &{\bf0.8911$\pm$.003}&{\bf0.8900$\pm$.003}&{\bf0.8916$\pm$.017}&{\bf0.8107$\pm$.018}&{\bf0.7995$\pm$.009}&{\bf0.8080$\pm$.010}\\
   	\cmidrule(r){1-1}\cmidrule(r){2-4}\cmidrule(r){5-7}
    %\sage w/ C=${|Y|}$ & 0.8801$\pm$0.0114 & 0.8582$\pm$0.0068 & 0.8692$\pm$0.0192 &{0.7823$\pm$0.0190}&{0.7913$\pm$0.0164}&{0.7949$\pm$0.0090}\\
		
   \sage w/ C=3&{0.8807$\pm$.015}&{0.8670$\pm$.005}&{0.8686$\pm$.004}&{0.7633$\pm$.008}&{0.7873$\pm$.014}&{0.7827$\pm$.017}\\
   	
   \sage w/ C=5&{0.8801$\pm$.014}&{0.8569$\pm$.012}&{0.8593$\pm$.011}&{\bf0.7927$\pm$.014}&{0.7904$\pm$.028}&{0.7886$\pm$.011}\\
			
   \sage w/ C=10&{0.8851$\pm$.003}&{0.8736$\pm$.022}&{0.8659$\pm$.015}&0.7769$\pm$.019&{\bf0.7940$\pm$.015}&{\bf0.8040$\pm$.022}\\
   \sage w/ C=15&{\bf0.8894$\pm$.016}&{\bf0.8883$\pm$.011}&{\bf0.8801$\pm$.009}&{0.7873$\pm$.012}&{0.7913$\pm$.021}&{0.7963$\pm$.015}\\
	
 \sage w/ C=20&{0.8883$\pm$.020}&0.8703$\pm$.007&0.8599$\pm$.009&0.7850$\pm$.008&0.7909$\pm$.022&0.7724$\pm$.018\\

    \toprule
   &\multicolumn{3}{c}{\textbf{PubMed}$\quad |Y|$=3}&\multicolumn{3}{c}{\textbf{MSAcademic}$\quad |Y|$=15}\\
   \cmidrule(r){1-1}\cmidrule(r){2-4}\cmidrule(r){5-7}
FedAvg&0.7149$\pm$.012&0.7260$\pm$.002&0.6954$\pm$.026&0.6986$
\pm$.002&{0.6908$\pm$.020}&{0.6705$\pm$.012}\\
\sage$_{\text{w/o \proto}}$&{\bf0.9090$\pm$.005}&{\bf0.8885$\pm$.002}&{\bf0.8697$\pm$.004}&{\bf0.9504$\pm$.004}&{\bf0.9455$\pm$.001}&{\bf0.9362$\pm$.001}\\
 	
		\cmidrule(r){1-1}\cmidrule(r){2-4}\cmidrule(r){5-7}
    %\sage w/ C=$|Y|$&{\bf0.9039$\pm$0.0069}&0.8862$\pm$0.0097&0.8650$\pm$0.0183&{\bf0.9452$\pm$0.0012}&{0.9393$\pm$0.0013}&{0.9302$\pm$0.0008}\\

\sage w/ C=3& {\bf0.8996$\pm$.007} & 0.8862$\pm$.010 & 0.8650$\pm$.018 & 0.9354$\pm$.001&{0.9365$\pm$.001}&{0.9314$\pm$.001}\\
   \sage w/ C=5&0.8929$\pm$.009&{\bf0.8872$\pm$.003}&{0.8652$\pm$.002}&{0.9353$\pm$.001}&{\bf0.9422$\pm$.001}&{\bf0.9351$\pm$.002}\\

   \sage w/ C=10&{0.8995$\pm$.005}&{0.8817$\pm$.005}&{0.8642$\pm$.008}&{0.9353$\pm$.001}&{0.9352$\pm$.001}&{0.9313$\pm$.001}\\
   \sage w/ C=15&{0.8961$\pm$.011}&{0.8804$\pm$.004}&{\bf0.8662$\pm$.003}&{\bf0.9452$\pm$.001}&{0.9393$\pm$.001}&{0.9302$\pm$.001}\\

   \sage w/ C=20 &0.8917$\pm$.004&0.8781$\pm$.006&0.8580$\pm$.007&0.9351$\pm$.001&{0.9353$\pm$.001}&{0.9313$\pm$.001}\\
    \bottomrule
  \end{tabular}
\end{table*}

\vspace{-2mm}
\subsection{Component Study of \gen}
\label{subsec:exp_gen}
\begin{figure}[t]
  \centering
\includegraphics[width=3in]{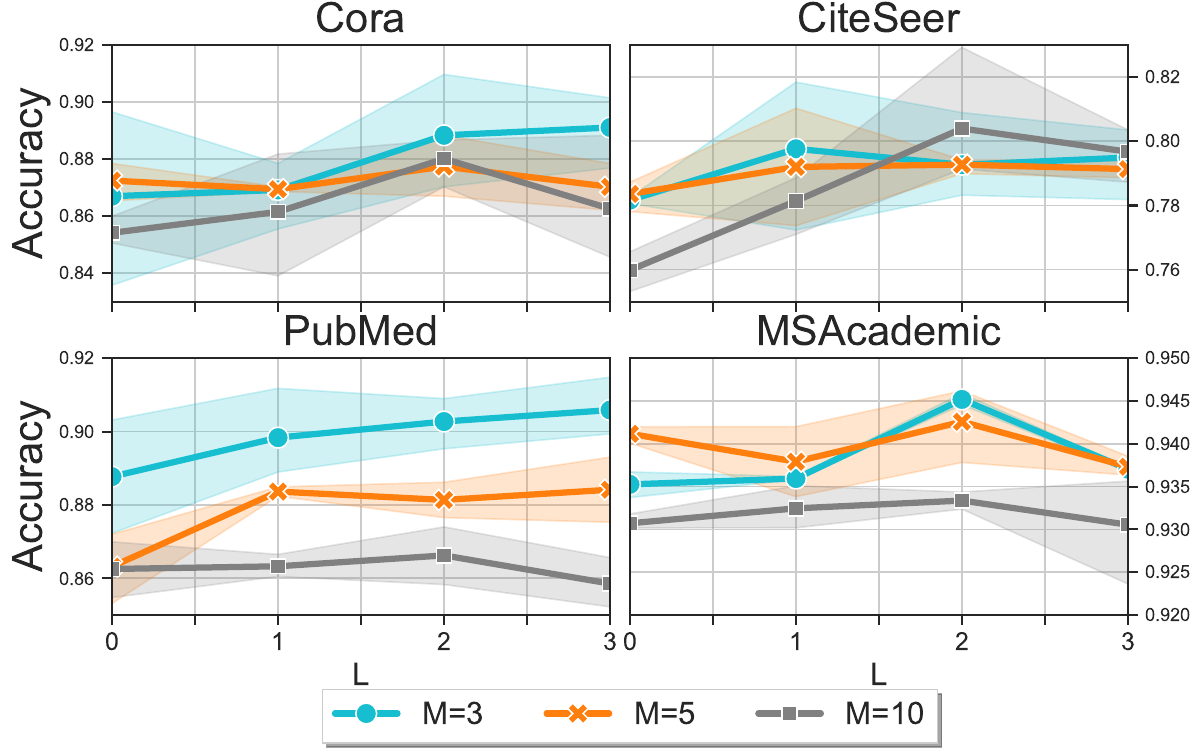}
\vspace*{-2mm}\caption{Component study for \gen in \sage with different depths $L$ of generated neighbor embeddings on four datasets with different $M$'s. $L$=0 is FedSage.\vspace{-5mm}} \label{fig:hops}
\end{figure}

We conduct in-depth studies for \gen with varying depth $L$ of the generated neighbors in \sage. As shown in Fig.~\ref{fig:hops}, 
% Specifically, we test different depth $L\in\{1,2,3\}$ on four datasets with different $M$s to study the influence of the deep neighbor depth on the outcome classifier. 
$L$ controls the amount of neighborhood information exchanged in the system. Positive $L$ can constantly elevate testing accuracy, compared with only exchanging neighbor features in FedSage ($L$=0). Across different datasets, the optimal $L$ is usually around 2. When a dataset has too many missing links between subgraphs (e.g., $M$=10), a large $L$ introduces more biased deep neighbor embeddings, and thus worse performances. 
% Interestingly, for the majority of tested scenarios, increasing $L$ does not result in obvious over-smoothing problem, \ie, steep accuracy decrease, as in the centralized GNN training~\cite{}.

\vspace{-1mm}
\subsection{Component Study of \proto}
\label{subsec:exp_proto}
\begin{SCfigure}
  \caption{Component study for \nfdp with different levels of edge-LDP privacy protection on MSAcademic with different client numbers.}\label{fig:dp}
 {\includegraphics[width=1.5in]{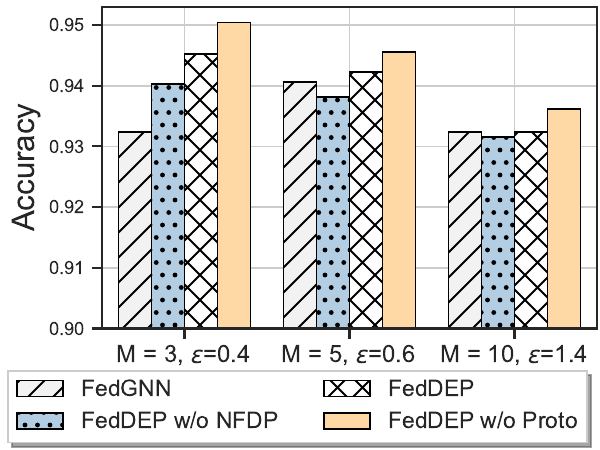}}% picture filename
\end{SCfigure}
We compare the downstream task performance of \sage under different numbers of cluster $C$ in \proto. 
Table~\ref{table:cluster} shows that choosing a proper $C$, which controls how abstract the exchanged information is within the system, can constantly elevate the final testing accuracy. Across different datasets, when $C$ is chosen around the number of classes, the performance is usually good. $C$ being too small like 3 or too large like 20 can result in slight performance drops, but \sage is in general insensitive to $C$ in a wide range. %For all datasets, , \eg, for $C$=15. %Across all datasets, with $C$=15, \sage improvements FedAvg by averaged 12.33\%.

% \begin{figure}[h]
% \centering
% \includegraphics[width=3.4in]{figures/box.pdf}
% \caption{Case study on the differences between models' soft predictions and the ground truth, on CiteSeer dataset, M=5.} \label{fig:box}
% \end{figure}

\vspace{-2mm}
\subsection{Component Study of \nfdp}
\label{subsec:exp_dp}
We compare models' utility under the same DP guarantees for noise-free frameworks (\sage and \sage$_{\text{w/o \proto}}$) and noise-injected frameworks (FedGNN and \sage$_{\text{w/o \nfdp}}$), as shown in Fig.~\ref{fig:dp}. 
% \todo{hiding $\delta$ values} 
% Specifically, when M=3, $d$=5, and $D$=15, we have (0.4,0.3)-edge-LDP, and $\sigma$ for FedGNN and \sage without NFDP is 4.16; when M=5, $d$=5, and $D$=10, we have (0.6,0.5)-edge-LDP, and $\sigma$ is 2.06; when M=10, $d$=5, and $D$=3, we have (1.4,0.9)-edge-LDP, and $\sigma$ is 0.38. 
In MSAcademic dataset, when $M$=3, we have $d$=5, $D$=15, $\varepsilon$=0.4, and $\sigma$=4.2; when $M$=5, we have $d$=5, $D$=10, $\varepsilon$=0.6, and $\sigma$=2.1; when $M$=10, we have $d$=5, $D$=3, $\varepsilon$=1.4, and $\sigma$ =0.4. Regarding Fig.~\ref{fig:dp}, our \nfdp always outperforms the noise-injected counterparts in achieving similar edge-LDP, which empirically justifies the superior utility-privacy trade-off of \nfdp when compared to gradients perturbation-based approaches such as DPSGD.

 \begin{figure}[t]
\centering
{\includegraphics[width=3.4in]{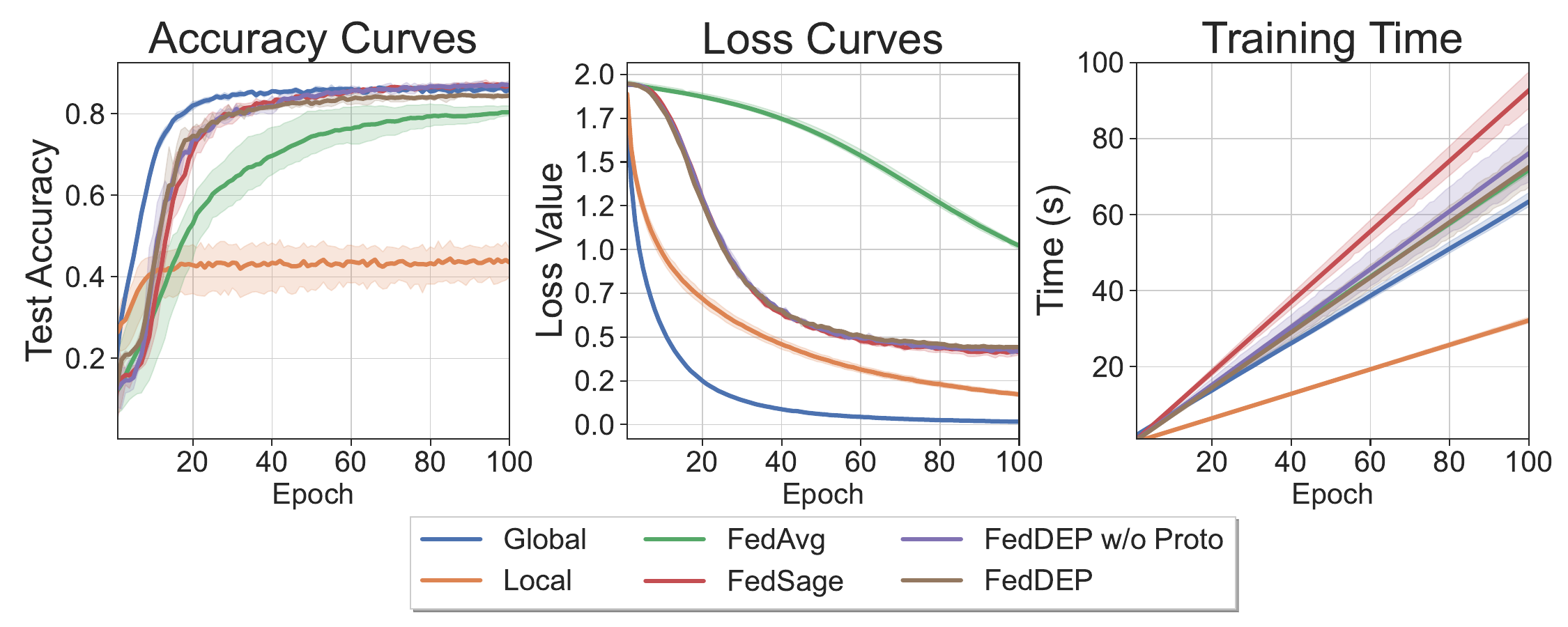}}
\vspace*{-3mm}\caption{Training curves of different frameworks on Cora dataset with M=5. (Best viewed in color.) \vspace*{-5mm}} \label{fig:curves}
\end{figure}
% We provide the case study with five clients on the CiteSeer dataset to inspect the differences between the ground truths (one-hot vectors) and the soft predictions of different models. As shown in Fig.~\ref{fig:box}, for all classes of CiteSeer, \sage and \sage w/o prototyping narrow the prediction gaps with the ground truths compared to the FedSage and local models, \ie, the mean value is much closer to zeros. The significant reduction further justifies the effectiveness of generating cross-subgraph missing deep neighbors. For half classes of CiteSeer, \sage and \sage w/o prototyping remarkably decrease the variance compared to FedSage, which validates the robustness of our proposed techniques.

\vspace{-2mm}
\subsection{Convergence Analysis}

For the Cora dataset with five data owners, we visualize testing accuracy, loss convergence, and runtime along 100 epochs in obtaining $\F$ with Global, Local, FedAvg, FedSage, \sage$_{\text{w/o \proto}}$, and \sage. The results are presented in Fig.~\ref{fig:curves}. Both \sage and \sage$_{\text{w/o \proto}}$ consistently achieve satisfactory convergence with rapidly improved testing accuracy. Regarding runtime, even though the classifiers from \sage$_{\text{w/o \proto}}$ and \sage learn from distributed mended subgraphs, they do not consume observable more training time compared to FedAvg. Compared to FedSage, \sage and its variation reduce the dimension of mended information, and thus save non-neglectable training time. Thanks to \proto, \sage consumes far less time than \sage$_{\text{w/o \proto}}$ and only slightly more time compared to Global and FedAvg.

%% file: 07-conclusion.tex
\section{Conclusion}

In this work, we study subgraph FL by comprehensively tackling the unique challenges %in subgraph FL that remains unsolved in pioneer works, \ie, 
of utility, efficiency, and privacy. We propose \sage, a novel subgraph FL framework with Deep Efficient Private neighbor generation. \sage consists of a series of techniques including deep neighbor generation with embedding-fused graph convolution, efficient pseudo-FL with missing neighbor prototyping, and privacy guarantee with noise-free edge-LDP. Theoretical analysis and empirical results together embrace the claimed benefits of \sage. 
%Future directions of \sage include the employment of more powerful GNNs such as based on graph transformers, pruning of unnecessary graph generation such as based on model uncertainty, stronger privacy protection such as node-LDP based on the noise-free DP and DPSGD, and applications on other real-world graph datasets.

\section*{Acknowledgement}
This research was partially supported by the collaborative research grant funding provided by the Halle Institute for Global Research at Emory University and the Office of International Cooperation and Exchanges at Nanjing University.

%% file: 08-for_pub_appendix.tex
\newpage

\appendix

\balance

\section*{A. Proof for Embedding-fused Graph Convolution}
\subsection*{A.1 Proof for Statement~3.1}
\begin{customst}{3.1} For a node $v$, at each layer of embedding-fused graph convolution, it aggregates nodes on the impaired ego-graph with the corresponding mended deep neighbor embeddings with separate learnable weights.
\end{customst}

\begin{proof}
 At $k$-th layer of embedding-fused graph convolution, for every node $u$ in $v$'s one-hop ego-graph $G^1(v)$, we denote its mean averaged node representations as $\bar{x}^k_u\in\mathbb{R}^{1\times d_h}$ and embeddings as $\bar{z}_u\in\mathbb{R}^{1\times d_z}$.

According to our description in Section~3, we have
\begin{equation*}
    x_u^k=\sigma(W^{(k)}\times [\bar{x}^{k-1}_u|| \bar{z}_u]^\top)^{\top},
\end{equation*}
where $W^{(k)}\in \mathbb{R}^{d_h\times (d_h+d_z)}$ is the learnable matrix in the convolution.

As $x_u^k$ can also be regarded as
\begin{equation*}
\small
   \sigma(\begin{gathered}
\begin{bmatrix} 
W^{x(k)}_{1,1}&...&W^{x(k)}_{1,d_h} & W^{z(k)}_{1,1}&...&W^{z(k)}_{1,d_z}\\
\vdots&...&\vdots&\vdots&...&\vdots\\
W^{x(k)}_{d_h,1}&...&W^{x(k)}_{d_h,d_h} & W^{z(k)}_{d_h,1}&...&W^{z(k)}_{d_h,d_z}\\
\end{bmatrix}\times
\begin{bmatrix} \bar{x}^{k-1}_{u,1}\\
\vdots\\
\bar{x}^{k-1}_{u,d_x}\\
\bar{z}_{u,1}\\
\vdots\\
\bar{z}_{u,d_z}\\
\end{bmatrix}
)^{\top},
\end{gathered}
\end{equation*}

which equals to $\sigma(W^{x(k)}\times {\bar{x}^{k-1\top}_u}+W^{z(k)}\times \bar{z}_u^\top)^{\top}$, where $W^{x(k)}\in \mathbb{R}^{d_h\times d_h}$ and $W^{z(k)}\in \mathbb{R}^{d_h\times d_z}$ are learnable weights in the convolution.

Therefore, we justify the correctness of embedding-fused graph convolution where the mended deep neighbors and the representations/features contribute to the convolution with respective learnable parameters, and conclude the proof.
\end{proof}

\subsection*{A.2 Proof for Statement~3.2}

\begin{customlemma}{A.1} For a node $v$, we denote the prediction, computed by one layer of embedding-fused graph convolution on its 1-hop impaired ego-graph, where every node is mended with deep neighbors computed on the respective $L$-hop missing context, as $\tilde{y}'_{v}$, and the prediction, computed by ($L$+1) layers of graph convolution on its ($L$+1)-hop ego-graph, as $\tilde{y}_v$, where $L\in \mathbb{N}^*$. $\tilde{y}'_v$ and $\tilde{y}_v$ are the compound vectors for the same local context of $v$.
\label{lemma:convolve-1-L}
\end{customlemma}

\begin{proof}

For node $v$, we compute its prediction $\tilde{y}'_{v}$ as 
\begin{equation*}
\begin{aligned}
    \tilde{y}'_{v}=x_v^1=&\sigma(W^{(1)}\times [ mean(\{x^0_u|u\in G^1(v)\})||\bar{z}_v]^\top,
    % =&\sigma(W^{z(0)}\times \bar{z}_v^\top+\\
    % &W^{x(1)}\times mean(\{\sigma(W^{x(0)}\times x_u^\top+W^{z(0)}\times \bar{z}_u^\top)|u\in\{\N(v)\cup v\}\})^\top)
    \end{aligned}
\end{equation*}
where for every $u\in G^1(v)$,
\begin{equation*}
\begin{aligned}
    x^0_u=&\sigma(W^{(0)}\times [x_u||\bar{z}_u])^\top=\sigma(W^{(0)}\times [x_u||mean(z_u)])^\top
    \end{aligned}
\end{equation*}

Since $x^0_u$ contains $\{x_u,z_u\}$, and $\tilde{y}'_{v}$ is then computed based on $\{x_u,z_u|u\in G^1(v)\}\cup\{\bar{z}_v\}$. We only need to verify $\{x_u,z_u|u\in G^1(v)\}\cup\{\bar{z}_v\}$ containing the same information as the $\{x_u|u\in G^{L+1}(v)\}$.

First we have $\{\bar{z}_v\}$ computed from the $L$-hop neighbors of $v$, \ie, $\{x_u|u\in G^{L}(v)\}$. Then we only need to consider whether the content of $\{x_u,z_u|u\in G^1(v)\}$ covers the $\{x_u|u\in G^{L+1}(v)\setminus G^{L}(v)\}$. Since every $z_u^p\in z_u$ is computed on the $L$-hop ego-graph of node $u$ with original graph convolution mechanism, $z_u^p$ contains the information of $\{x_p|p\in G^L(u)\}$. Thus, the union of $z_u$ for $u\in G^1(v)$ covers $\{x_p|p\in G^{L}(u),u\in G^1(v)\}=\{x_p|p\in G^{L+1}(v)\}$, which includes $\{x_u|u\in G^{L+1}(v)\setminus G^{L}(v)\}$.

Obviously, $\{x_u,z_u|u\in G^1(v)\}\cup\{\bar{z}_v\}$ contains the same $L+1$ ego-graph content as $\{x_u|u\in G^{L+1}(v)\}$ does, we have Lemma~\ref{lemma:convolve-1-L} proved.
\end{proof}

\begin{customst}{3.2} For a node $v$, we denote the prediction, computed by $K$ layers of embedding-fused graph convolution on its $K$-hop impaired ego-graph mended with deep neighbors of $L$-hop local contexts, as $\tilde{y}'_v$, and the prediction, computed by ($K$+$L$) layers of graph convolution on its ($K$+$L$)-hop ego-graph, as $\tilde{y}_v$, where $K,L\in \mathbb{N}^*$. $\tilde{y}'_v$ and $\tilde{y}_v$ are the compound vectors for the same local context of $v$.
\end{customst}

\begin{proof}
To prove Statement~3.2, we extend Lemma~\ref{lemma:convolve-1-L} from 1-hop impaired ego-graph to the $K$-hop impaired ego-graph mended with $L$-hop local missing context embeddings. 

By iterativly applying Lemma~\ref{lemma:convolve-1-L} $K$-$L$ times, we have node $v$'s prediction $\tilde{y}'_v$ computed on $\{x_u,z_u|u\in G^{K}(v)\}$ with $z_u$ containing the information of $\{x_p|p\in G^{L}(u)\}$. The entire content is the same as where $\tilde{y}_v$ is retrieved with original graph convolution, \ie, $\{x_p|p\in G^{K+L}(u)\}$. In this way, we have Statement~3.2 proved.

\end{proof}

\section*{B. Proof for Theorem~3.1}
\begin{customlemma}{B.1} Given a graph, with its nodes' degrees by at least $D$, and a GCN model for embedding computation, after one epoch of mini-batch training on 1-hop ego-graphs drawn from the graph with sampling size as $d$, the GCN achieves at most $(\ln\frac{D+1}{D+1-d}, \frac{d}{D})$-edge-LDP when $d< D$, and at least $(d\ln\frac{D+1}{D}, 1-(\frac{D-1}{D})^d)$-edge-LDP otherwise.
\label{lemma:edge-ldp-1-1}
\end{customlemma}

\begin{proof} To prove Lemma~\ref{lemma:edge-ldp-1-1}, we first revisit the NFDP mechanisms \cite{ijcai2021p216} on $(\varepsilon, \delta)$-differential privacy of different sampling policies.

\begin{customthm}{B.1}[NFDP mechanism-I~\cite{ijcai2021p216}]
Given a training dataset of
size $D$, sampling without replacement achieves $(\ln \frac{D+1}{D+1-d}, \frac{d}{D})$-
differential privacy, where $d$ is the subsample size.
\end{customthm}

\begin{customthm}{B.2}[NFDP mechanism-II~\cite{ijcai2021p216}]
Given a training dataset of
size $D$, sampling with replacement achieves $(d\ln\frac{D+1}{D}, 1-(\frac{D-1}{D})^d)$-
differential privacy, where $d$ is the subsample size.
\end{customthm}

To apply Theorem B.1 and Theorem B.2 in Lemma~\ref{lemma:edge-ldp-1-1}, we can regard the 1-hop neighbors of the target node $v$, \ie, the neighbors on the 1-hop ego-graph of $v$, as the entire dataset with size $D$, and the mini-batch sampling node size is the subsampling size $d$.

In this way, one epoch of training the GCN model with the mini-batch sampling has two cases. One case is when $d< D$, while the other is $d\geq D$. For the neighbor sampling method, we follow the implementation of FederatedScope~\cite{wang2022federatedscope}, where the former case uses the sampling without replacement, and the latter case uses the sampling with replacement. Therefore, when $d< D$, the sampling can achieve $(\ln\frac{D+1}{D+1-d}, \frac{d}{D})$-differential privacy for the neighbor list, and $(d\ln\frac{D+1}{D}, 1-(\frac{D-1}{D})^d)$-differential privacy otherwise.

To transfer the general DP to the edge-LDP, we need to analyze it according to the definition of edge-LDP and differential privacy. We revisit the definition of general DP as follows.

\begin{customdef}{B.1}[$(\varepsilon, \delta)$-differential privacy]
A randomized mechanism $\mathcal{M}: \mathcal{A} \rightarrow B$ with domain $\mathcal{A}$ and range B satisfies $(\varepsilon, \delta)$-differential privacy if for all two neighboring inputs $U, U'\in \mathcal{A}$ that differ by one record, and any measurable subset of outputs $S \subseteq B$ it holds that
\begin{equation}
Pr[\mathcal{M}(U)\in S]\leq e^{\varepsilon}Pr[\mathcal{M}(U')\in S]+\delta
\label{eq:dp}
\end{equation}
\end{customdef}

Then we revisit the definition of edge-LDP as below.
\begin{customdef}
{B.2} For a graph with $n$ nodes, denote its node $v$’s neighbor list as $(b_1,\dots,b_n)$. For $u\in[n]$, if $v$ is linked with $v$, $b_u$ is 1. Otherwise, $b_u$ is 1.
Let $\varepsilon, \delta \in \mathbb{R}_{\geq 0}$, and $R: \mathcal{G} \rightarrow \mathbb{R}$ is a randomized algorithm. $R$ provides $(\varepsilon,\delta)$-edge-LDP if for any two local neighbor lists $\gamma, \gamma'$ that differ in one bit and any $S \subseteq R$,
\begin{equation}
    Pr[R(\gamma) \in S] \leq e^{\varepsilon} Pr[R(\gamma') \in S]+\delta.
    \label{eq:edge-ldp}
\end{equation}
\end{customdef}

By regarding the input dataset $U, U'$ in Eq.~\eqref{eq:dp} as two neighbor lists $\gamma, \gamma'$ in Eq.~\eqref{eq:edge-ldp}, we have general differential privacy transferred to edge-LDP. As the mini-batch sampling GCN can achieve $\gamma, \gamma'$ in Eq.~\eqref{eq:edge-ldp} through whether sampling a neighbor node in the ego-graph, we transfer the sampling in NFDP of $(\varepsilon, \delta)$-differential privacy to the equal effect of the mini-batching sampling in noise-free $(\varepsilon, \delta)$-edge-LDP.

Since nodes on a graph can have different degrees, and the lower bound of protection implies the privacy of this mechanism, we choose the max values of $(\varepsilon, \delta)$ by calculating them using the minimum degree among all nodes. In this way, Lemma~\ref{lemma:edge-ldp-1-1} is proved.
\end{proof}

\begin{customlemma}{B.2} For a subgraph, given every node's $L$-hop ego-graph with its every $L$-1 hop nodes of degrees by at least $D$, and a GCN model for embedding computation, after $N$ epochs of mini-batch training with each hop of sampling size as $d$, the GCN achieves $(\tilde{\varepsilon},\tilde{\delta})$-edge-LDP, where
{\small
\begin{equation*}
\begin{aligned}
\tilde{\varepsilon}&=\min \{LN \varepsilon, LN \varepsilon\frac{(e^{\varepsilon}-1)}{e^{\varepsilon}+1}+\varepsilon U \sqrt{2 LN}\},\\
\tilde{\delta}&=(1-\delta)^{LN}(1-\delta'),
\end{aligned}
\end{equation*}
}
and $U = \min \{\sqrt{\ln (e+\frac{\varepsilon\sqrt{LN }}{\delta'})}, \sqrt{\ln (\frac{1}{\delta'})}\}$, for $\delta'\in[0,1]$, and $(\varepsilon,\delta)$ are $(\ln\frac{D+1}{D+1-d}, \frac{d}{D})$ and $(d\ln\frac{D+1}{D}, 1-(\frac{D-1}{D})^d)$ in Lemma~\ref{lemma:edge-ldp-1-1} for respective cases.
\label{lemma:edge-ldp-N-L}
\end{customlemma}

\begin{proof}
To prove Lemma~\ref{lemma:edge-ldp-N-L}, we need to adaptively apply Lemma~\ref{lemma:edge-ldp-1-1} by $N$ epochs on the $L$ times of graph convolution, \ie, total $LN$ times. Thus, we revisit the Composition of DP Mechanisms~\cite{kairouz2015composition} as follows.

\begin{customthm}{B.3}[Composition of DP~\cite{kairouz2015composition}]  For any $\varepsilon>0$, $\delta, \delta' \in[0,1]>0$, the class of $(\varepsilon,\delta)$-differential private mechanisms satisfies $(\tilde{\varepsilon},1-(1-\delta)^k(1-\delta'))$-differential private under $k$-fold adaptive composition, for 
{\scriptsize
\begin{equation*}
\begin{aligned}
\tilde{\varepsilon}&=\min \{k \varepsilon, k \varepsilon\frac{(e^{\varepsilon}-1)}{e^{\varepsilon}+1}+\varepsilon\sqrt{2 k}\min \{\sqrt{\ln (e+\frac{\varepsilon\sqrt{k }}{\delta'})},
% , LN\varepsilon \frac{\left(e^{\varepsilon}-1\right) }{e^{\varepsilon}+1}+\varepsilon 
\sqrt{\ln (\frac{1}{\delta'})}\}\}
\end{aligned}
\end{equation*}
}
\end{customthm}

By firstly aligning general differential privacy to edge-LDP as we described in the proof of Lemma~\ref{lemma:edge-ldp-1-1}, obviously, we have the same conclusion of the composition rule for edge-LDP as Theorem B.3. Then we substitute the $k$ in the composition rule to $LN$, and specifying the $(\epsilon,\delta)$ as the pairs in Lemma~\ref{lemma:edge-ldp-1-1}. Thus, Lemma~\ref{lemma:edge-ldp-N-L} is proved.
\end{proof}

\begin{customthm}{3.1}[Noise-free edge-LDP of \sage] 
For a distributed subgraph system, on each subgraph, given every node's $L$-hop ego-graph with its every $L$-1 hop neighbors of degrees by at least $D$, \sage unifies all subgraphs in the system to federally train a joint model of a classifier and a cross-subgraph deep neighbor generator. By learning from deep neighbor embeddings that are obtained from locally trained GNNs in $N$ epochs of mini-batch training with a sampling size for each hop as $d$, \sage achieves $(\log(1+r(e^{\tilde{\varepsilon}}\text{-}1),r\tilde{\delta})$-edge-LDP, where
{\small
\begin{equation*}
\begin{aligned}
\tilde{\varepsilon}&=\min \{LN \varepsilon, LN \varepsilon\frac{(e^{\varepsilon}-1)}{e^{\varepsilon}+1}+\varepsilon U\sqrt{2 LN} \},\\
\tilde{\delta}&=(1-\delta)^{LN}(1-\delta'), \quad \delta'\in[0,1],
\end{aligned}
\end{equation*}
}
and $U = \min \{\sqrt{\ln (e+\frac{\varepsilon\sqrt{LN }}{\delta'})}, \sqrt{\ln (\frac{1}{\delta'})}\}$. $r$ is the expected value of the Bernoulli sampler in \gen. When $d<D$, $(\varepsilon,\delta)$ are tighter than $(\ln\frac{D+1}{D+1-d}, \frac{d}{D})$; when $d\geq D$, $(\varepsilon,\delta)$ are tighter than $(d\ln\frac{D+1}{D}, 1-(\frac{D-1}{D})^d)$. Both pairs of $(\varepsilon,\delta)$ serve as the lower bounds of the edge-LDP protection under the corresponding cases.
\end{customthm}

\begin{proof}

\sage framework first pre-calculates the embeddings from a mini-batch trained GCN to retrieve prototype sets, then it leverages the deep neighbor generator that employs a Bernoulli sampler $R$ with expected value $r$ to jointly train a classifier on subgraphs mended with generated deep neighbor prototypes. 

To prove Theorem~3.1, we revisit the privacy amplification by subsampling in the general DP~\cite{balle2020privacy}.

\begin{customthm}{B.4}[privacy amplification~\cite{balle2020privacy}] Given a dataset $U$ with $n$ data records, subsampling mechanism $\mathcal{S}$ subsamples a subset of data $\{d_i|\sigma_i=1,i\in[n]\}$ by sampling $\sigma_i\sim Ber(p)$ independently for $i\in[n]$. If mechanism $\mathcal{M}$ satisfied $(\varepsilon,\delta)$-differential privacy, mechanism $\mathcal{M}\circ\mathcal{S}$ is $(\log(1+p(e^{\varepsilon-1}),p\delta)$-differential private.
\end{customthm}

We prove Theorem~3.1 by applying Theorem~B.4 and Lemma~\ref{lemma:edge-ldp-N-L} in four steps.

We first transfer the conclusion of Theorem B.4 into edge-LDP by following the proof of Lemma~\ref{lemma:edge-ldp-1-1}. Then we specify the $(\varepsilon,\delta)$-differential privacy mechanism $\mathcal{M}$ in Theorem~B.4 as the edge-LDP embedding computation GCN model in Lemma~\ref{lemma:edge-ldp-N-L} with respective privacy-related parameters. Next, we specify the subsampling mechanism $\mathcal{S}$ in Theorem~B.4 as the Bernoulli sampler in \sage with \gen on prototypes. By substituting the $p$ in Theorem~B.4 to $r$, we have Theorem~3.1 proved.

\end{proof}

\section*{C. Related Works}
\subsection*{C.1 Federated Learning for Graphs}
\label{subsec:rw_fl}
With massive graph data separately stored by distributed data owners, recent research has emerged in the field of FL over graph data. Some studies propose FL methods for tasks on distributed knowledge graphs, such as recommendation or representation learning~\cite{peng2021differentially, chen2021fede, zhang2022efficient, gu2023dynamic}. Another direction is for the scenarios where every client holds a set of small graphs, such as molecular graphs for drug discovery~\cite{xie2021federated}. 
In this work, we consider subgraph FL, where each client holds a subgraph of the entire global graph, and the only central server is dataless. The instrumental isolation of data samples leads to incomplete structural features of local nodes due to cross-subgraph neighbors missing not at random, which is fundamentally different from the centralized graph learning scenarios with unbiased sparse links \cite{liu2022local} or randomized DropEdge \cite{rong2019dropedge}. 

To deal with the missing neighbor problem in subgraph FL, existing works~\cite{zhang2021subgraph,zhang2022subgraph,wu2022federated,chen2021fedgraph, pan2023lumos, zhao2022fedgsl} propose to augment local subgraphs by retrieving missing neighbors across clients, and then mend the subgraphs with the retrieved neighbor information. FedGraph~\cite{chen2021fedgraph} considers a relaxed scenario where the existences of inter-subgraph neighbors are known for corresponding clients. Lumos~\cite{pan2023lumos}, as well as FedGraph~\cite{chen2021fedgraph}, requests the central server to manage the FL process with auxiliary data. FedSage~\cite{zhang2021subgraph} primarily focuses on the design of the missing neighbor generator without considering the important aspects of efficiency and privacy. FedHG~\cite{zhang2022subgraph} studies the heterogeneous subgraph FL systems where graphs consist of multiple types of nodes and links, and it only protects the partial privacy of certain types of nodes in the system. FedGNN~\cite{wu2022federated} and FedGSL~\cite{zhao2022fedgsl} equip their augmentation with privacy protection based on %. However, compared to the system we consider, 
additional trusted authorities and/or noise injection. % is required by FedGNN to achieve its protection. 

None of them provides a complete solution to the utility, efficiency, and privacy of subgraph FL. %All these works are deficient in efficiency as their augmentations induce substantial overhead in computation and communication. 

%Our proposed subgraph FL framework is built for subgraph FL systems and comprehensively tackles three key challenges in subgraph FL, \ie, utility, efficiency, and privacy.
\subsection*{C.2 Privacy-Preserving Learning for Graphs}
\label{subsec:rw_dp}
Privacy-preserving learning over graph data has been widely studied. Differential Privacy (DP)~\cite{dwork2006differential} is a widely applied privacy concept in this field, which describes the privacy of a method in protecting individual samples while preserving the analytical properties of the entire dataset.
A prevalent approach in attaining a graph mining model with general DP is DPSGD~\cite{abadi2016deep}, which injects designed noise into clipped gradients during model training. For centralized training scenarios, DPGGAN~\cite{yang2021secure} incorporates DPSGD to achieve DP for individual links on original graphs.
In FL systems, VFGNN~\cite{zhou2021vertically} and FedGNN~\cite{wu2022federated} leverage DPSGD and cryptology techniques to obtain rigorous privacy guarantees for federated graph learning. Meanwhile, to achieve general DP on graphs, there are some other noised-injecting based methods. Previous works of centralized learning ~\cite{lu2014exponential,ahmed2019publishing,xiao2014differentially}, FKGE~\cite{peng2021differentially} and FedGSL~\cite{zhao2022fedgsl} for FL systems, guarantee their proposed techniques with general DP by applying noise perturbation. 
 %In FL systems, FKGE~\cite{peng2021differentially} is a DP framework for learning embedding from distributed knowledge graphs. Specifically, FKGE perturbs the outputs of distributed models to achieve general DP.
 %However, VFGNN studies the vertical FL system, where each client holds a fraction of features for their nodes. Though FedGNN focuses on distributed subgraphs, homomorphic encryption can result in significant additional cost for the FL system. 
 
%However, applying noise that perturbs gradients or/and outputs can significantly degenerate the final model performance.
However, general DP does not depict the protections for sensitive node features, edges, or neighborhoods, on distributed graphs. %Yet VFGNN and FedGNN only analyzes its general differential privacy for the entire system, without showing the protection for sensitive neighborhood structures of nodes in a distributed graph.
Edge local DP and node local DP (edge-LDP and node-LDP) are two specific types of DP targeting local nodes' neighbor lists~\cite{qin2017generating}. These novel DP definitions better fit the graph learning that learns from multiple neighbor lists, and match the privacy goal of protecting nodes' local neighborhoods. 

As illustrated in Definition 2.2 in~\cite{qin2017generating}, edge-LDP defines how much a model tells for two neighborhoods that differ by one edge, while node-LDP promises a model's max leakage for all possible neighborhoods. In contrast to node-LDP, which is much stronger and can severely hinder the graph model's utility, edge-LDP precisely illustrates the local DP for local neighborhoods without overly constraining the model.

% \vspace{-4pt}
% \begin{definition} [Edge Local Differential Privacy \cite{qin2017generating,liu2022collecting}]
% \label{def:edge-ldp}
% For a graph with $n$ nodes, denote its node $v$’s neighbor list as $(b_1,\dots,b_n)$. For $u\in[n]$, if $v$ is linked with \red{$u$}, $b_u$ is 1. Otherwise, $b_u$ is 0.
% Let $\varepsilon, \delta \in \mathbb{R}_{\geq 0}$, and $R: \mathcal{G} \rightarrow \mathbb{R}$ is a randomized algorithm. $R$ provides $(\varepsilon,\delta)$-edge-LDP if for any two local neighbor lists $\gamma, \gamma'$ that differ in one bit and any $S \subseteq R$,
% \vspace{-1mm}
% \begin{equation*}
% \small
%     Pr[R(\gamma) \in S] \leq e^{\varepsilon} Pr[R(\gamma') \in S]+\delta.
% \end{equation*}
% \end{definition}
% \vspace{-2mm}

There are several works analyzing edge-LDP over distributed graph data. Qin et al.~\cite{qin2017generating} propose a decentralized social graphs generation technique with the edge-LDP. Imola et al.~\cite{imola2022differentially} analyze the edge-LDP of the proposed shuffle techniques in handling the triangle and 4-cycle counting for neighbor lists of distributed users. Lin et al.~\cite{lin2022towards} propose Solitude, an edge-LDP collaborative training framework for distributed graphs, where each client shares its perturbed local graph for the training. However, different from our subgraph FL setting, its central server (data curator) has access to node identities and labels. To the best of our knowledge, we are the first to leverage edge-LDP in the FL setting.

\begin{table*}[t]
\centering
\footnotesize
\caption{Datasets and the synthesized distributed systems statistics. $|V_i|$ and $|E_i|$ rows show the averaged numbers of nodes and links in all subgraphs, and $\Delta E$ shows the total number of missing cross-subgraph links.}
  \label{table:data-table}
\begin{tabular}{crrrrrrrrrrrr}
   \toprule
Dataset & \multicolumn{3}{c}{Cora} & \multicolumn{3}{c}{Citeseer} & \multicolumn{3}{c}{PubMed} & \multicolumn{3}{c}{MSAcademic} \\
\midrule
$(|V|, |E|)$ &\multicolumn{3}{c}{(2708, 5278)} & \multicolumn{3}{c}{(3327, 4552)} & \multicolumn{3}{c}{(19717, 44324)} & \multicolumn{3}{c}{(18333, 81894)} \\
$(d_x,|Y|)$& \multicolumn{3}{c}{(1433, 7)} & \multicolumn{3}{c}{(3703, 6)} & \multicolumn{3}{c}{(500, 3)} & \multicolumn{3}{c}{(6805, 15)} \\
\midrule
M&\multicolumn{1}{c}{3}&\multicolumn{1}{c}{5}&\multicolumn{1}{c}{10}&\multicolumn{1}{c}{3}&\multicolumn{1}{c}{5}&\multicolumn{1}{c}{10}&\multicolumn{1}{c}{3}&\multicolumn{1}{c}{5}&\multicolumn{1}{c}{10}&\multicolumn{1}{c}{3}&\multicolumn{1}{c}{5}&\multicolumn{1}{c}{10}\\
\cmidrule(rl){2-4}\cmidrule(rl){5-7}\cmidrule(rl){8-10}\cmidrule(rl){11-13}
$|V_i|$&{903}&  {542}&    {271}&    {1109} &   {665}   &  {333} & 6572 &  3943   &  1972  & 6111 & 3667  &  1833   \\
$|E_i|$& 1594 &  945 &  437 & 1458 &   866 &  431   & 13251 &7901     &   3500  & 24300  & 13949  &  5492  \\
$\Delta E$& 496 &  552  &  912   & 178     & 224    & 247 &   4570 & 4818    &9323 &  8995    &12149  & 26973\\
$\Delta E/|E|$&0.0940 & 0.1046& 0.1728& 0.0391& 0.0492& 0.0543& 0.1031&0.1087&0.2103&0.1098&0.1484&0.3294\\
\bottomrule
\end{tabular}
\end{table*}

\section*{D. Revisit of FedSage+}
\label{subsec:fedsage+}

In this section, we revisit the popular existing subgraph federated learning framework, i.e., FedSage+, the variant of FedSage with the proposed missing neighbor generator (NeighGen) \cite{zhang2021subgraph}. %in , which studies the federated node classification problem over a distributed subgraph system as described in Section~\ref{subsec:problem}. 
For simplicity, in this paper, we refer to this stronger variant as FedSage.

\subsection*{D.1 Neighbor Generation} 
The proposed NeighGen in \cite{zhang2021subgraph} includes an encoder $H^{e}$ and a generator $H^{g}$.
% , which are parameterized by $\theta^e$ and $\{\theta^d,\theta^f\}$ respectively.
For a node $v$ on $G_i$, NeighGen generates its missing neighbors by taking in its $K$-hop ego-graph $G_i^K(v)$. Specifically, it predicts the number of $v$'s missing neighbors $\tilde{n}_v$
% by $\tilde{n}_v= \sigma(\theta^{d\top}\cdot \he(G_i^K(v);\theta^e))$
, and predicts the respective feature set $\tilde{x}_v$.

\subsection*{D.2 Cross-subgraph~Neighbor~Reconstruction} 

To obtain ground truth for supervising NeighGen without actually seeing the missing neighbors, each client simulates the missing neighbor situation by randomly holding out a pre-determined portion of the nodes and all links involving them. To allow a NeighGen model to generate diverse and realistic missing neighbors, the system conducts federated cross-subgraph training as follows.%neighbor reconstruction loss for nodes in $\bar{V}_i$ is computed as follows.

\begin{enumerate}[leftmargin=*]
    \item Each client $D_i$ sends its local NeighGen's generator $H^{g}$ and its input to all other clients $D_j$.
    \item $D_j$ computes the cross-subgraph feature reconstruction loss $\Lf_{i,j}$ between real node features on $G_j$ and the generated ones from received data. 
%     as follows %Specifically, it computes 
%     \vspace{-5pt}
% {\small
% \begin{equation*}
% \begin{aligned}
% \label{eq:fedsage_fed}
% \Lf_{i,j} = \frac{1}{|\bar{V}_i|} \sum_{v\in \bar{V}_i} \sum_{p\in[\tilde{n}_v]}
% 	\min_{u\in V_j}
% 	(||H^{g}(h^K_v)^{p}-x_{u}||^2_2).
% % 	(||\tilde{x}_{v}^{p}-x^{q}||^2_2)
% \end{aligned}
% \end{equation*}
% }\vspace{-5pt}
%where $H_{g}(h^K_v)^{p}$ is the $p$-th predicted feature.
\item $D_j$ sends $\Lf_{i,j}$'s gradients back to $D_i$ via server $S$.

\item $D_i$ computes the total gradients of cross-subgraph neighbor reconstruction loss $\Lf_{i}=\alpha^n\sum_{j\in[M]}\Lf_{i,j}$ by summing up all received gradients from other clients. Notably, $\Lf_{i,i}$ is the local neighbor reconstruction loss computed on local ground truth obtained from hidden nodes and edges.
% , that is
% \vspace{-5pt}
% {\small
% \begin{equation*}
% \label{eq:local_neighgen}
% \begin{aligned}
%  \vspace{-2pt}\Lf_{i,i}= \frac{1}{|\bar{V}_i|} \sum_{v\in \bar{V}_i}[\alpha^d L_1^S(\tilde{n}_v-n_v)
%     +\alpha^f \sum_{p\in[\tilde{n}_v]}
% 	\min_{u\in \bar{\mathcal{N}}_{i}}(||\tilde{x}_{v}^{p}-x_{u}||^2_2)],\vspace{-2pt}
% \end{aligned}
% \end{equation*}}
% \vspace{-5pt}
%where $L_1^S$ is the smooth L1 distance~\cite{girshick2015fast}, $\bar{\mathcal{N}}_{i}$ contains $v$'s neighbors on $G_i$ that are missed into $V_i^h$, and $\alpha$s are hyper-parameters.
\end{enumerate}

To attain the generalized final classifier, in FedSage, data owners federally train a shared model of NeighGen with a GraphSage classifier, where the classifier learns on nodes drawn from local subgraphs mended with the generated neighbors. For more technical details of the process and equations, please refer to the original paper of FedSage~\cite{zhang2021subgraph}.

\section*{E. Additional Experimental Details}
We present the statistics of tested four datasets and the synthesized distributed systems in Tab.~\ref{table:data-table}.